\documentclass[10pt,letterpaper]{article}
\usepackage{copyro-preprint}
\newcommand{\method}{\textsc{CoPyro}}
\newcommand{\blk}{B}            
\newcommand{\nslot}{K}          
\newcommand{\msk}{\texttt{[M]}} 
\newcommand{\Emb}{\mathbf{E}}   

\newcommand{\papertitle}{CoPyro: Feeding Copy Proposals to Parallel Drafters}
\hypersetup{pdftitle={\papertitle},pdfauthor={Hong-Kai Zheng, Piji Li},
  pdfsubject={Technical Report},pdfkeywords={Machine Learning, ICML}}

\begin{document}

\twocolumn[{%
  \paperfront
    {\textcolor{pine}{CoPyro:} Feeding Copy Proposals\\to Parallel Drafters}
    {{\sffamily\bfseries\fontsize{11}{13}\selectfont
      Hong-Kai Zheng\textsuperscript{1}\par
      Piji Li\textsuperscript{1*}\par}
     \vspace{6pt}
     {\sffamily\fontsize{8.6}{10.7}\selectfont\color{secondary}
       \textsuperscript{1}Nanjing University of Aeronautics~and~Astronautics\par
       \vspace{5pt}
       \textsuperscript{*}corresponding author\par
       \vspace{5pt}
       Correspondence to:\par
       Hong-Kai Zheng \textless{}\href{mailto:dt_ttt@nuaa.edu.cn}{dt\_ttt@nuaa.edu.cn}\textgreater{},\par
       Piji Li \textless{}\href{mailto:pjli@nuaa.edu.cn}{pjli@nuaa.edu.cn}\textgreater{}.\par
       \vspace{5pt}
       The code will be released at:\par
       \href{https://github.com/dt-3t/CoPyro}{github.com/dt-3t/CoPyro}.\par}}
    {Speculative decoding accelerates language generation by letting a target model verify
several draft tokens in one forward pass.
Earlier neural drafters such as EAGLE extend each candidate sequence one token at a time.
Recent work has turned to parallel drafters such as DFlash, which predict a whole block at once.
Another approach uses $n$-gram retrieval to find an earlier occurrence of the current
context suffix and copy the tokens that followed it directly into the draft.
We introduce \method{}, which feeds these copied tokens into a parallel drafter in place
of masks. The drafter learns from context which tokens to keep and which to revise
as it predicts the block in its usual forward pass.
Across eight benchmarks on Qwen3-4B, \method{} improves overall acceptance length
for all four parallel drafters tested, with gains of 2.19--5.08\% when trained on 100K samples.
With DFlash trained on 710K samples, overall gains range from 1.08\% to 2.00\%
across the tested model sizes and temperatures. Code generation benefits most in both settings.
On Qwen3-8B, processing the proposals adds an average of 0.24\,ms per decoding round in Transformers.
}%
}]

\section{Introduction}
\label{sec:introduction}

Generating a long response with an autoregressive language model requires a separate
target-model evaluation for each new token. Speculative decoding reduces this cost by
drafting several tokens cheaply and verifying them together in one target-model
pass~\citep{leviathan2023fast,chen2023accelerating}. Neural drafters such as EAGLE
extend each candidate sequence one token at a time, so drafting a longer sequence
requires more sequential steps~\citep{li2024eagle}. More recent parallel drafters
such as DFlash predict a whole block in one forward pass~\citep{chen2026dflash}.
DFlash conditions its predictions on features from the target model, but each future
position enters the drafter as a mask. Could the context supply a cheap first guess for some of these positions?

An existing source of such guesses is $n$-gram retrieval. When the current suffix has
appeared earlier in the context, the tokens that followed it can be copied and submitted
directly to the target model as a draft~\citep{saxena2023pld,hu2024sam}. A copied
continuation can be useful even when it is only partly correct. Consider two calls to
the same function that take different input tensors but share the same keyword
arguments. If retrieval copies the wrong tensor name, verification stops at the first
mismatched token. The keyword arguments that follow may still be exactly what the
current call needs.

Hybrid methods such as SAM-Decoding use match length to choose between a retrieved
continuation and a neural draft~\citep{hu2024sam}. In the function-call example, that
choice applies to the entire continuation, although its usefulness varies from token to
token. Giving the parallel drafter the copied tokens as inputs would let it use the
surrounding code to predict the appropriate tensor name while retaining the shared
arguments.

We introduce \method{}, which feeds retrieved continuations to a parallel drafter
during both training and inference. Before each drafting step, \method{} replaces
the masks at the proposed positions with embeddings of the copied tokens. Two learned
embeddings tell the drafter how long the suffix match was and how often other matches
support each proposed token. The drafter then learns which tokens to keep and which
to revise in the same forward pass that predicts the block. During training, we
randomly drop proposals so the drafter also learns to predict from masks when no
proposal is available. The target model verifies the resulting block as usual.
Figure~\ref{fig:overview} illustrates one decoding round.

We evaluate \method{} on eight benchmarks spanning math, code, and dialogue. On
Qwen3-4B, we train DFlash, Domino~\citep{huang2026domino}, DSpark~\citep{dspark2026},
and DFlash2~\citep{dflash2} on 100K samples. Adding \method{} improves overall
acceptance length by 2.19--5.08\% and end-to-end
throughput by 0.99--4.04\% over their respective baselines. Code generation benefits
most. On HumanEval, for example, acceptance length increases by 5.67--9.03\%
across the four drafters. The gains are smaller with DFlash trained on 710K samples,
where overall throughput improves by 0.39--1.33\% across the tested Qwen3-4B and
Qwen3-8B configurations. In a controlled Qwen3-8B measurement with Transformers,
processing the proposals adds an average of 0.24\,ms per decoding round.

Our contributions are threefold.
\begin{itemize}
  \item We introduce \method{}, which lets parallel drafters learn to use retrieved
    continuations as inputs.
  \item We design lightweight match-length and agreement embeddings that help the
    drafter assess individual copied tokens within its usual forward pass.
  \item We evaluate \method{} across four parallel drafters and analyze when the
    drafter corrects copied tokens, how much even short matches contribute, and how
    much proposal processing costs.
\end{itemize}

\section{Related Work}
\label{sec:related}

\paragraph{Model-based speculative decoding.}
Speculative decoding drafts several tokens cheaply and verifies them with a single forward pass
of the target model~\citep{leviathan2023fast,chen2023accelerating}.
Autoregressive drafters generate tokens sequentially along each
candidate path~\citep{li2024eagle,li2024eagle2,li2025eagle3,cheng2024redrafter,
zhang2024draftverify,elhoushi2024layerskip,zhou2024distillspec,ankner2024hydra,deepseek2024v3}.
Building on blockwise decoding~\citep{stern2018blockwise}, parallel drafters predict a block
in one pass~\citep{cai2024medusa,chen2026dflash,an2025pard,monea2023pass,hui2026peagle,bhendawade2025streaming,
li2026diffuspec,xiao2024parallelspec,gao2025falcon,lin2024bita}.
To model dependencies within a block, some drafters add a lightweight sequential head
after the parallel backbone or combine its logits with those of a short-context model~\citep{huang2026domino,dspark2026,zheng2026dels}.
Others score candidate pairs at adjacent positions~\citep{dflash2,eisenstadt2026dcoco}
or inject causal information at each drafter
layer~\citep{zhang2026dspine}. Other methods refine the block after drafting~\citep{xpress2026,yu2026dedit}.
\citet{kim2024refinement} rescore drafts with $n$-gram or neural language models.
DART uses $n$-gram probabilities to prune its draft tree~\citep{liu2026dart}.
ReTrace conditions DFlash on hidden representations of the previous round's rejected
suffix~\citep{lin2026retrace}.
DDTree constructs a draft tree from DFlash's per-position
distributions~\citep{ringel2026ddtree}.
DMTP uses pretrained MTP to enhance first-position prediction~\citep{shen2026dmtp}.
For mask-based drafters, \method{} complements these designs by changing only the
draft inputs.

\paragraph{Model-free speculative decoding.}
Model-free methods typically draft by retrieving the continuation of a matched pattern: from the
prompt or reference documents~\citep{yang2023llma,saxena2023pld}, from an external
datastore~\citep{he2024rest}, from an $n$-gram pool populated by Jacobi iterations~\citep{fu2024lookahead}, from a suffix
automaton~\citep{hu2024sam} or suffix trees over previous
requests~\citep{oliaro2025suffixdecoding}, or from a token adjacency
matrix~\citep{luo2025tokenrecycling}.
Copying from the context is inexpensive and works especially well when the output repeats
the input~\citep{saxena2023pld,ranjan2026hybridspark}.
Hybrid methods combine retrieval with model-based drafting by choosing between draft
sources~\citep{hu2024sam,su2026hybridverified,ranjan2026hybridspark}, merging retrieved
candidates into a draft tree~\citep{quan2025rasd,graft2026}, extending drafts with retrieved
phrases~\citep{zhao2024ouroboros}, or using retrieval to accelerate the draft model itself~\citep{shen2026double}.
In all of these, retrieved tokens serve as drafts that are verified as they are; \method{}
instead feeds them to a neural drafter as input and lets it learn how to use them.

\paragraph{Proposals and refinement.}
Two-stage object detectors first generate cheap region proposals and then classify and refine
them with a learned network~\citep{girshick2014rcnn,ren2015faster}.
In text generation, copy mechanisms let a model copy tokens from its input~\citep{vinyals2015pointer,
gu2016copynet}, retrieve-and-edit methods generate by editing a retrieved
example~\citep{guu2018editing,hashimoto2018retrieve}, and non-autoregressive models iteratively
refine a masked draft~\citep{ghazvininejad2019maskpredict}.
In non-autoregressive translation, TM-LevT learns to edit a translation retrieved from
memory~\citep{xu2023tmlevt}.
\method{} brings the proposal-then-refine idea to speculative drafting: a copied proposal is
refined by the drafter in the same single forward pass that would otherwise start from masks.

\section{Preliminaries}
\label{sec:preliminaries}

Let $p$ be the target model and $x_{1:a}$ the sequence committed so far, where the last token
$x_a$, the \emph{anchor}, was sampled by the target model in the previous round.
A DFlash-style drafter~\citep{chen2026dflash} with block size $\blk$ proposes the next
$\blk-1$ tokens in parallel.
Its input is a block of $\blk$ embeddings placed at positions $a, a+1, \dots, a+\blk-1$,
\begin{equation}
  \mathbf{z}^{(0)} = \bigl(\Emb(x_a),\; \Emb(\msk),\; \dots,\; \Emb(\msk)\bigr),
  \label{eq:plain-block}
\end{equation}
where $\Emb$ is the frozen input embedding of the target model and $\msk$ a mask token.
The drafter attends to this block bidirectionally and to features of the target model's hidden
states over the context, and an output head turns the resulting block states into draft tokens
$\hat y_{1:\blk-1}$.
The target model scores the anchor and all draft tokens in one forward pass, accepts the longest
prefix of the draft that agrees with its own samples, and appends one token of its own; that token
becomes the anchor of the next round.
Each round therefore commits between $1$ and $\blk$ tokens.

\section{\method{}}
\label{sec:copy_proposals}

\begin{figure*}[t]
  \centering
  \includegraphics[width=0.97\textwidth]{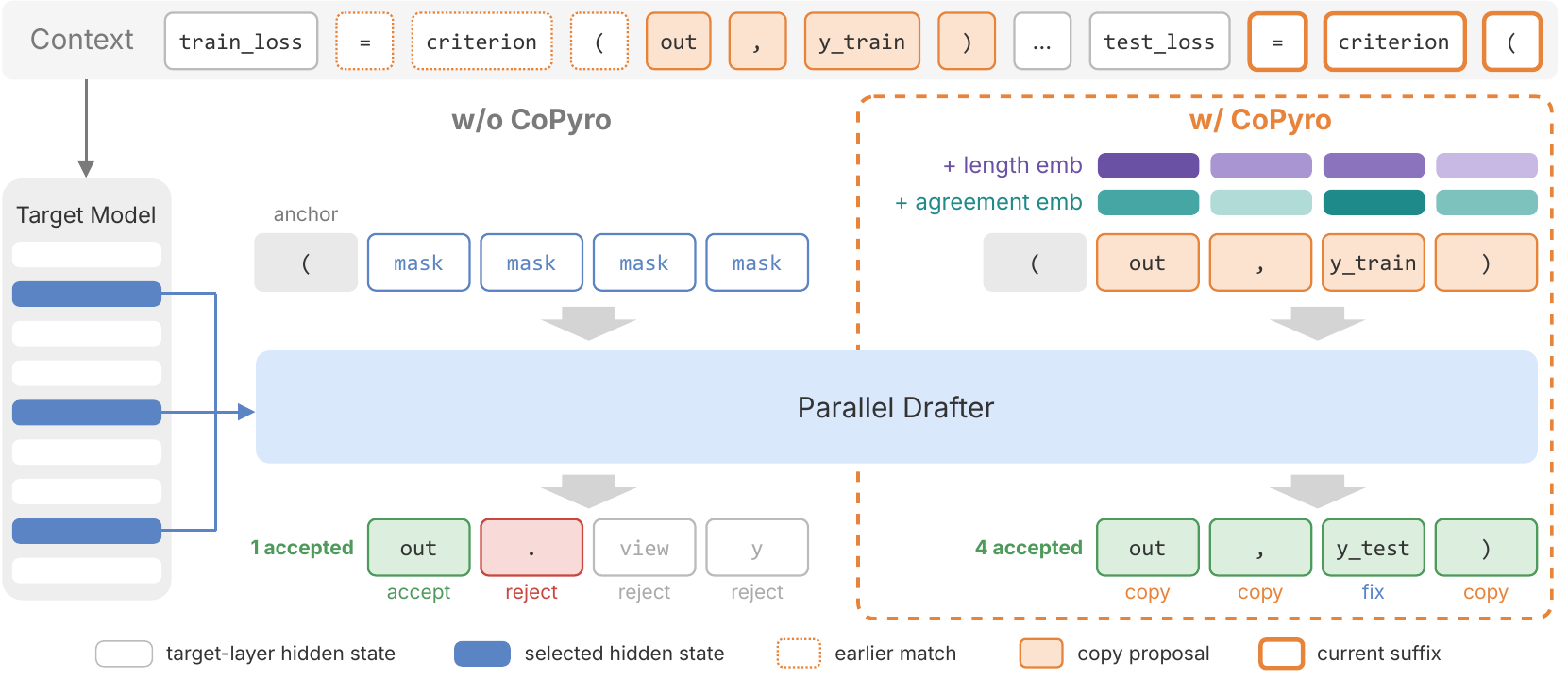}
  \caption{Overview of \method{}. Before each drafting step, \method{} finds the longest suffix of
    the context that occurred earlier (dotted) and copies the tokens that followed it as a proposal.
    The proposal replaces the mask inputs of the parallel drafter, together with learned
    match-length and agreement embeddings; the drafter, its output head and the target model's
    verification are unchanged.}
  \label{fig:overview}
\end{figure*}

Two-stage object detectors generate candidate regions and use a learned network to classify and
refine them~\citep{girshick2014rcnn,ren2015faster}.
Inspired by this, we give the drafter in speculative decoding a \emph{proposal}: a preliminary
but cheap guess of the upcoming tokens.
Parallel drafters are a natural fit for such input: draft positions see one another and the
context.
Through training, the drafter learns how to use the proposal, for example keeping it, correcting
it, drawing on it when drafting other positions, or simply ignoring it.

\subsection{Retrieval}
\label{sec:method:retrieval}

\method{} obtains proposals by matching $n$-grams in the context, as in model-free
speculative decoding~\citep{saxena2023pld,hu2024sam}.
At the start of each decoding round, it finds the longest suffix of $x_{1:a}$ with an earlier
occurrence that has a usable continuation.
If the selected occurrence ends at $j^\star < a$, the length-$L$ match satisfies
\begin{equation}
  S_L = x_{a-L+1:a} = x_{j^\star-L+1:j^\star}.
\end{equation}
We copy the available continuation, up to $\blk-1$ tokens, to obtain the proposal
\begin{equation}
  c_{1:\nslot} = x_{j^\star+1:j^\star+\nslot},
  \label{eq:retrieval-proposal}
\end{equation}
where $\nslot$ is the number of copied tokens.
If no proposal is found, $\nslot=0$ and the drafter uses the plain block~\eqref{eq:plain-block}.
Appendix~\ref{app:retrieval} gives the full retrieval procedure.

\subsection{Reliability features}
\label{sec:method:features}

We design two features to help the drafter learn how to use $n$-gram proposals of varying quality.

\paragraph{Match length.}
Longer matches are more specific.
We map the match length $L$ to one of $N_{\mathrm{len}}$ buckets, $b(L)$, with one bucket per
length for short matches and progressively wider buckets for long ones.

\paragraph{Agreement.}
Earlier occurrences of a pattern can have different continuations.
Let $\mathcal{O} = \{\, i < a : x_{i-L+1:i} = S_L \,\}$ be the end positions of the earlier
occurrences of the matched suffix, and $n = |\mathcal{O}|$.
For $k = 1, \dots, \nslot$, let
\begin{equation}
  n_k = \bigl|\{\, i \in \mathcal{O} : i+k \le a,\ x_{i+1:i+k} = c_{1:k} \,\}\bigr|
\end{equation}
be the number of these occurrences whose continuation, within the committed sequence, matches the
first $k$ proposal tokens.
Both counts include the selected occurrence $j^\star$, so $1 \le n_k \le n$.
The agreement bucket at slot $k$ is
\begin{equation}
  g_k =
  \begin{cases}
    0, & n = 1,\\
    \operatorname{Bucket}_{\mathrm{agr}}(\alpha_k), & n > 1,
  \end{cases}
  \qquad
  \alpha_k = \frac{n_k - 1}{n - 1},
\end{equation}
where $\alpha_k$ is the fraction of the other occurrences, excluding $j^\star$, that
agree with the proposal up to slot $k$.
Appendix~\ref{app:feature-buckets} describes both bucket mappings.

\subsection{Feeding proposals to the drafter}
\label{sec:method:input}

A proposal can be given to the drafter in many ways; a natural one is to substitute it for the
masks.
The masks carry no information, so replacing them changes neither the drafter's architecture nor
its computational cost.
Concretely, \method{} replaces the first $\nslot$ mask embeddings of the block with the embeddings
of the proposed tokens and adds a match-length embedding and an agreement embedding:
\begin{equation}
  \mathbf{z}_k =
  \begin{cases}
    \Emb(x_a) & k = 0,\\[2pt]
    \Emb(c_k) + \mathbf{T}^{\mathrm{len}}_{k,\, b(L)} + \mathbf{T}^{\mathrm{agr}}_{k,\, g_k}
      & 1 \le k \le \nslot,\\[2pt]
    \Emb(\msk) & k > \nslot,
  \end{cases}
  \label{eq:copy-block}
\end{equation}
where $\mathbf{T}^{\mathrm{len}} \in \mathbb{R}^{(\blk-1)\times N_{\mathrm{len}}\times d}$ and
$\mathbf{T}^{\mathrm{agr}} \in \mathbb{R}^{(\blk-1)\times N_{\mathrm{agr}}\times d}$.
Here $N_{\mathrm{len}}$ and $N_{\mathrm{agr}}$ are the numbers of length and agreement
buckets, respectively, and $d$ is the hidden size.
Indexing the tables by slot lets the same feature value mean different things at different
distances from the anchor.

The rest of the drafter, including its output head and its alignment between inputs and
predictions, is unchanged.
Because the proposal is an \emph{input} rather than a draft, the drafter learns how to use it,
and whether to use it at all.
Verification is also unchanged.

\subsection{Training and Inference}
\label{sec:method:train-infer}

\paragraph{Training.}

Each drafter keeps its own training objective; only its input block changes from
\eqref{eq:plain-block} to \eqref{eq:copy-block}.
At every anchor $a$, the proposal and its features are computed with the retrieval rule of
\S\ref{sec:method:retrieval} from the prefix $x_{1:a}$ alone, so they never contain the tokens
being predicted.
With probability $p$, we drop the whole proposal of a block, so that the drafter stays
competent without proposals.

\paragraph{Inference.}

The retrieval index is built over the prompt and the first target token during prefill.
After each verification, the tokens committed in that round (the accepted draft tokens and the
target's own token) are appended to the index, and the proposal for the next round is queried.
We measure the added cost of this processing in Section~\ref{sec:exp:latency}.
Figure~\ref{fig:overview} illustrates one decoding round of \method{}.

\newcommand{\exppending}{\textemdash}
\newcommand{\exppair}{\exppending\,/\,\exppending}
\newcommand{\expplotplaceholder}[3]{%
  \fbox{\parbox[c][#1][c]{\dimexpr\linewidth-2\fboxsep-2\fboxrule\relax}{%
    \centering\small #2\\[0.5em]
    \footnotesize #3\\[0.5em]\textit{Results pending}}}}

\section{Experiments}
\label{sec:exp}

\begin{table*}[t]
  \centering
  \caption{Main results after one epoch on 710K PerfectBlend samples.
    Cells give $\tau\,/\,S_{\mathrm{AR}}$: acceptance length and end-to-end speedup over AR, including prefill.
    Overall pools all 994 requests within each target/temperature configuration.
    Bold marks the highest displayed value for each metric among all four methods within each model/temperature block.}
  \label{tab:exp:main}
  \begingroup
  \small
  \setlength{\tabcolsep}{2pt}
  \renewcommand{\arraystretch}{1.04}
  \resizebox{\textwidth}{!}{%
  \begin{tabular}{l*{9}{c}}
    \toprule
    \multirow{2}{*}{Method} & \multicolumn{3}{c}{\textsc{Math}} & \multicolumn{3}{c}{\textsc{Code}} & \multicolumn{2}{c}{\textsc{Chat}} & \textsc{Overall} \\
    \cmidrule(lr){2-4}\cmidrule(lr){5-7}\cmidrule(lr){8-9}\cmidrule(lr){10-10}
    & GSM8K & MATH-500 & AIME25 & HumanEval & MBPP & LCB & MT-Bench & Alpaca & \textit{Avg.} \\
    \midrule
    \multicolumn{10}{l}{\textbf{Qwen3-4B, $T=0$}} \\
    DFlash & $\mathbf{7.67}\,/\,\mathbf{5.45}$ & $7.37\,/\,\mathbf{5.12}$ & $5.94\,/\,\mathbf{4.38}$ & $5.30\,/\,3.95$ & $5.02\,/\,3.63$ & $5.42\,/\,4.11$ & $4.07\,/\,2.41$ & $3.06\,/\,2.14$ & $5.38\,/\,3.55$ \\
    Pure $n$-gram & $1.13\,/\,1.14$ & $1.38\,/\,1.44$ & $1.64\,/\,1.55$ & $1.38\,/\,1.36$ & $1.10\,/\,1.13$ & $1.29\,/\,1.44$ & $1.25\,/\,1.14$ & $1.09\,/\,1.09$ & $1.25\,/\,1.29$ \\
    DFlash + Switch & $7.08\,/\,5.11$ & $6.57\,/\,4.70$ & $5.47\,/\,4.19$ & $5.30\,/\,3.99$ & $4.92\,/\,3.59$ & $5.49\,/\,\mathbf{4.27}$ & $3.90\,/\,2.37$ & $3.04\,/\,2.15$ & $5.16\,/\,3.48$ \\
    DFlash + \method{} & $7.63\,/\,5.38$ & $\mathbf{7.39}\,/\,\mathbf{5.12}$ & $\mathbf{5.97}\,/\,\mathbf{4.38}$ & $\mathbf{5.49}\,/\,\mathbf{4.05}$ & $\mathbf{5.04}\,/\,\mathbf{3.64}$ & $\mathbf{5.54}\,/\,4.19$ & $\mathbf{4.14}\,/\,\mathbf{2.42}$ & $\mathbf{3.11}\,/\,\mathbf{2.17}$ & $\mathbf{5.45}\,/\,\mathbf{3.58}$ \\
    \midrule
    \multicolumn{10}{l}{\textbf{Qwen3-4B, $T=1$}} \\
    DFlash & $6.92\,/\,4.70$ & $\mathbf{6.19}\,/\,3.83$ & $\mathbf{4.23}\,/\,\mathbf{2.96}$ & $4.94\,/\,3.60$ & $4.69\,/\,3.29$ & $4.94\,/\,3.07$ & $3.76\,/\,\mathbf{2.20}$ & $2.91\,/\,\mathbf{1.99}$ & $4.85\,/\,3.00$ \\
    Pure $n$-gram & $1.13\,/\,1.13$ & $1.30\,/\,1.32$ & $1.28\,/\,1.25$ & $1.32\,/\,1.29$ & $1.07\,/\,1.09$ & $1.17\,/\,1.22$ & $1.23\,/\,1.11$ & $1.05\,/\,1.04$ & $1.19\,/\,1.19$ \\
    DFlash + Switch & $6.41\,/\,4.56$ & $5.53\,/\,3.68$ & $3.90\,/\,2.88$ & $4.97\,/\,3.65$ & $4.58\,/\,3.25$ & $4.84\,/\,3.06$ & $3.61\,/\,2.17$ & $2.87\,/\,\mathbf{1.99}$ & $4.64\,/\,2.96$ \\
    DFlash + \method{} & $\mathbf{6.93}\,/\,\mathbf{4.79}$ & $6.13\,/\,\mathbf{3.84}$ & $4.11\,/\,2.93$ & $\mathbf{5.12}\,/\,\mathbf{3.74}$ & $\mathbf{4.74}\,/\,\mathbf{3.31}$ & $\mathbf{5.01}\,/\,\mathbf{3.16}$ & $\mathbf{3.83}\,/\,\mathbf{2.20}$ & $\mathbf{2.95}\,/\,\mathbf{1.99}$ & $\mathbf{4.90}\,/\,\mathbf{3.02}$ \\
    \midrule
    \multicolumn{10}{l}{\textbf{Qwen3-8B, $T=0$}} \\
    DFlash & $\mathbf{7.70}\,/\,\mathbf{5.34}$ & $7.48\,/\,5.13$ & $5.80\,/\,4.24$ & $5.34\,/\,3.96$ & $4.95\,/\,3.64$ & $5.53\,/\,4.25$ & $4.02\,/\,2.41$ & $3.08\,/\,\mathbf{2.06}$ & $5.40\,/\,3.54$ \\
    Pure $n$-gram & $1.14\,/\,1.13$ & $1.37\,/\,1.40$ & $1.48\,/\,1.45$ & $1.39\,/\,1.36$ & $1.08\,/\,1.10$ & $1.37\,/\,1.52$ & $1.19\,/\,1.13$ & $1.05\,/\,1.06$ & $1.24\,/\,1.27$ \\
    DFlash + Switch & $7.07\,/\,4.97$ & $6.62\,/\,4.63$ & $5.28\,/\,3.98$ & $5.31\,/\,3.97$ & $4.86\,/\,3.57$ & $5.60\,/\,4.35$ & $3.82\,/\,2.34$ & $3.04\,/\,2.04$ & $5.15\,/\,3.43$ \\
    DFlash + \method{} & $7.67\,/\,5.32$ & $\mathbf{7.56}\,/\,\mathbf{5.21}$ & $\mathbf{5.91}\,/\,\mathbf{4.31}$ & $\mathbf{5.54}\,/\,\mathbf{4.07}$ & $\mathbf{5.00}\,/\,\mathbf{3.67}$ & $\mathbf{5.69}\,/\,\mathbf{4.36}$ & $\mathbf{4.10}\,/\,\mathbf{2.43}$ & $\mathbf{3.10}\,/\,\mathbf{2.06}$ & $\mathbf{5.49}\,/\,\mathbf{3.58}$ \\
    \midrule
    \multicolumn{10}{l}{\textbf{Qwen3-8B, $T=1$}} \\
    DFlash & $\mathbf{6.80}\,/\,\mathbf{4.61}$ & $\mathbf{6.00}\,/\,\mathbf{3.91}$ & $\mathbf{4.19}\,/\,\mathbf{3.03}$ & $4.65\,/\,3.40$ & $4.44\,/\,3.17$ & $4.85\,/\,2.93$ & $3.55\,/\,2.10$ & $2.91\,/\,\mathbf{1.93}$ & $4.68\,/\,2.92$ \\
    Pure $n$-gram & $1.14\,/\,1.14$ & $1.32\,/\,1.34$ & $1.33\,/\,1.30$ & $1.33\,/\,1.31$ & $1.08\,/\,1.08$ & $1.13\,/\,1.13$ & $1.17\,/\,1.11$ & $1.04\,/\,1.03$ & $1.18\,/\,1.19$ \\
    DFlash + Switch & $6.34\,/\,4.36$ & $5.42\,/\,3.57$ & $3.80\,/\,2.80$ & $4.64\,/\,3.40$ & $4.41\,/\,3.12$ & $4.92\,/\,2.97$ & $3.39\,/\,2.08$ & $2.85\,/\,1.91$ & $4.51\,/\,2.84$ \\
    DFlash + \method{} & $6.78\,/\,4.57$ & $5.98\,/\,3.84$ & $4.15\,/\,2.95$ & $\mathbf{4.76}\,/\,\mathbf{3.46}$ & $\mathbf{4.58}\,/\,\mathbf{3.19}$ & $\mathbf{5.17}\,/\,\mathbf{3.12}$ & $\mathbf{3.68}\,/\,\mathbf{2.12}$ & $\mathbf{2.92}\,/\,1.89$ & $\mathbf{4.78}\,/\,\mathbf{2.93}$ \\
    \bottomrule
  \end{tabular}%
  }
  \endgroup
\end{table*}

\begin{table*}[t]
  \centering
  \caption{Generality across eight benchmarks on Qwen3-4B after one epoch on
    Qwen3-4B-Instruct-100K, $T=0$. Each cell reports $\tau\,/\,S_{\mathrm{AR}}$:
    acceptance length and the reported end-to-end throughput ratio to independently measured AR,
    including prefill. Overall pools all benchmarks. Bold marks the higher displayed
    value within each drafter pair.}
  \label{tab:exp:generality}
  \begingroup
  \small
  \setlength{\tabcolsep}{3pt}
  \renewcommand{\arraystretch}{1.08}
  \resizebox{\textwidth}{!}{%
  \begin{tabular}{l*{9}{c}}
    \toprule
    Drafter & GSM8K & MATH-500 & AIME25 & HumanEval & MBPP & LCB & MT-Bench & Alpaca & Overall \\
    \midrule
    DFlash & $4.33\,/\,3.16\times$ & $5.13\,/\,3.77\times$ & $4.47\,/\,3.36\times$ & $4.30\,/\,3.18\times$ & $4.02\,/\,3.01\times$ & $4.14\,/\,3.16\times$ & $3.07\,/\,1.98\times$ & $2.38\,/\,1.73\times$ & $3.91\,/\,2.76\times$ \\
    DFlash + \method{} & $\mathbf{4.36}\,/\,\mathbf{3.18}\times$ & $\mathbf{5.26}\,/\,\mathbf{3.87}\times$ & $\mathbf{4.71}\,/\,\mathbf{3.53}\times$ & $\mathbf{4.68}\,/\,\mathbf{3.40}\times$ & $\mathbf{4.17}\,/\,\mathbf{3.11}\times$ & $\mathbf{4.50}\,/\,\mathbf{3.44}\times$ & $\mathbf{3.27}\,/\,\mathbf{2.06}\times$ & $\mathbf{2.46}\,/\,\mathbf{1.77}\times$ & $\mathbf{4.11}\,/\,\mathbf{2.87}\times$ \\
    \midrule
    Domino & $4.99\,/\,2.79\times$ & $6.15\,/\,3.43\times$ & $5.32\,/\,3.05\times$ & $5.06\,/\,2.82\times$ & $4.84\,/\,2.73\times$ & $4.87\,/\,2.81\times$ & $3.52\,/\,1.68\times$ & $2.64\,/\,1.44\times$ & $4.59\,/\,2.42\times$ \\
    Domino + \method{} & $\mathbf{5.05}\,/\,\mathbf{2.80}\times$ & $\mathbf{6.25}\,/\,\mathbf{3.48}\times$ & $\mathbf{5.49}\,/\,\mathbf{3.13}\times$ & $\mathbf{5.42}\,/\,\mathbf{2.98}\times$ & $\mathbf{4.95}\,/\,\mathbf{2.79}\times$ & $\mathbf{5.17}\,/\,\mathbf{2.96}\times$ & $\mathbf{3.60}\,/\,\mathbf{1.73}\times$ & $\mathbf{2.68}\,/\,\mathbf{1.45}\times$ & $\mathbf{4.74}\,/\,\mathbf{2.48}\times$ \\
    \midrule
    DSpark & $\mathbf{4.92}\,/\,\mathbf{3.50}\times$ & $6.06\,/\,4.30\times$ & $5.19\,/\,3.79\times$ & $4.89\,/\,3.48\times$ & $\mathbf{4.69}\,/\,3.37\times$ & $4.73\,/\,3.47\times$ & $3.42\,/\,2.10\times$ & $2.58\,/\,1.77\times$ & $4.47\,/\,3.00\times$ \\
    DSpark + \method{} & $4.84\,/\,3.41\times$ & $\mathbf{6.08}\,/\,4.30\times$ & $\mathbf{5.34}\,/\,\mathbf{3.87}\times$ & $\mathbf{5.17}\,/\,\mathbf{3.63}\times$ & $4.68\,/\,3.37\times$ & $\mathbf{5.05}\,/\,\mathbf{3.66}\times$ & $\mathbf{3.49}\,/\,\mathbf{2.11}\times$ & $\mathbf{2.61}\,/\,1.77\times$ & $\mathbf{4.57}\,/\,\mathbf{3.03}\times$ \\
    \midrule
    DFlash2 & $5.13\,/\,3.32\times$ & $6.30\,/\,4.08\times$ & $5.42\,/\,3.60\times$ & $5.21\,/\,3.37\times$ & $4.88\,/\,3.19\times$ & $4.99\,/\,3.35\times$ & $3.60\,/\,2.01\times$ & $2.67\,/\,1.70\times$ & $4.69\,/\,2.87\times$ \\
    DFlash2 + \method{} & $\mathbf{5.19}\,/\,\mathbf{3.35}\times$ & $\mathbf{6.38}\,/\,\mathbf{4.12}\times$ & $\mathbf{5.54}\,/\,\mathbf{3.66}\times$ & $\mathbf{5.52}\,/\,\mathbf{3.53}\times$ & $\mathbf{4.99}\,/\,\mathbf{3.26}\times$ & $\mathbf{5.19}\,/\,\mathbf{3.46}\times$ & $\mathbf{3.73}\,/\,\mathbf{2.03}\times$ & $\mathbf{2.72}\,/\,\mathbf{1.71}\times$ & $\mathbf{4.83}\,/\,\mathbf{2.93}\times$ \\
    \bottomrule
  \end{tabular}%
  }
  \endgroup
\end{table*}

\begin{table*}[t]
  \centering
  \caption{Component ablations on DFlash with Qwen3-4B, one epoch on Qwen3-4B-Instruct-100K,
    and $T=0$. Entries are acceptance lengths $\tau$.
    Proposal variants use $m=1$ and $p=0.2$. Bold marks the highest displayed value in each column.}
  \label{tab:exp:components}
  \begingroup
  \small
  \setlength{\tabcolsep}{3pt}
  \renewcommand{\arraystretch}{1.08}
  \resizebox{\textwidth}{!}{%
  \begin{tabular}{l*{9}{c}}
    \toprule
    Variant & GSM8K & MATH-500 & AIME25 & HumanEval & MBPP & LCB & MT-Bench & Alpaca & Overall \\
    \midrule
    DFlash & 4.325 & 5.134 & 4.475 & 4.296 & 4.019 & 4.141 & 3.075 & 2.381 & 3.914 \\
    + copy proposal & 4.331 & 5.210 & 4.659 & 4.632 & 4.099 & 4.424 & 3.229 & 2.420 & 4.062 \\
    + proposal + length emb. & 4.357 & 5.263 & 4.676 & 4.682 & 4.121 & \textbf{4.505} & 3.247 & 2.455 & 4.102 \\
    + proposal + agreement emb. & \textbf{4.368} & \textbf{5.265} & 4.698 & 4.668 & 4.147 & 4.457 & 3.249 & 2.437 & 4.097 \\
    Full \method{} & 4.355 & 5.262 & \textbf{4.709} & \textbf{4.684} & \textbf{4.171} & 4.503 & \textbf{3.267} & \textbf{2.461} & \textbf{4.113} \\
    \bottomrule
  \end{tabular}%
  }
  \endgroup
\end{table*}

\begin{figure*}[t]
  \centering
  \captionsetup{skip=4pt}
  \captionsetup[subfigure]{skip=2pt}
  \begin{subfigure}[t]{0.40\textwidth}
    \includegraphics[width=\linewidth]{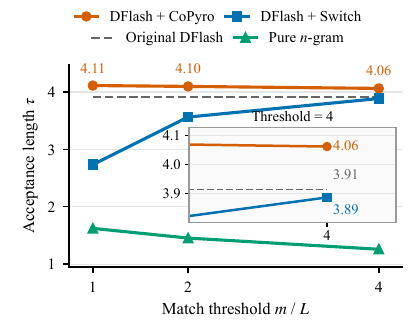}
    \caption{Overall.}
    \label{fig:exp:match-sweep}
  \end{subfigure}\hfill
  \begin{subfigure}[t]{0.32\textwidth}
    \includegraphics[width=\linewidth]{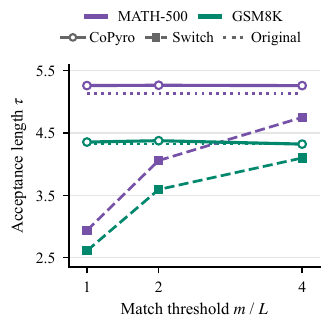}
    \caption{MATH-500 and GSM8K.}
    \label{fig:exp:math-sweep}
  \end{subfigure}\hfill
  \begin{subfigure}[t]{0.24\textwidth}
    \includegraphics[width=\linewidth]{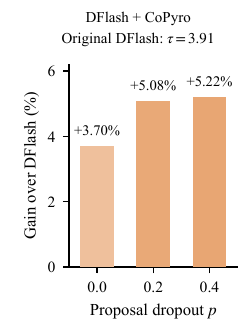}
    \caption{Proposal dropout.}
    \label{fig:exp:dropout-sweep}
  \end{subfigure}
  \caption{Design ablations on Qwen3-4B with DFlash at $T=0$.
    (a,b) Sweeps of minimum match length $m$ for \method{} and pure $n$-gram,
    and switching threshold $L$ for Switch ($p=0.2$).
    Horizontal lines show the original DFlash references; the inset magnifies threshold 4.
    (c) Proposal dropout at $m=1$, with acceptance-length gains measured relative to the original DFlash.}
  \label{fig:exp:ablations}
\end{figure*}

\begin{table*}[t]
  \centering
  \caption{Concurrent serving on Qwen3-8B with SGLang on HumanEval and LCB at $T\in\{0,1\}$.
    Columns are client concurrency; cells give mean TPS($\tau$) over four runs.
    TPS counts output tokens/s; $\tau$ is mean per-request acceptance length.
    Bold marks the larger displayed value for each metric within each DFlash--\method{} pair.}
  \label{tab:exp:concurrency}
  \begingroup
  \small
  \setlength{\tabcolsep}{4pt}
  \renewcommand{\arraystretch}{1.05}
  \begin{tabular*}{\textwidth}{@{\extracolsep{\fill}}cl*{6}{c}@{}}
    \toprule
    Bench. & Method & 1 & 2 & 4 & 8 & 16 & 32 \\
    \midrule
    \multirow{2}{*}{\shortstack{HumanEval\\$(T=0)$}} & DFlash & $161.9(5.28)$ & $283.8(5.31)$ & $\mathbf{478.2}(5.29)$ & $639.7(5.33)$ & $624.2(5.28)$ & $618.4(5.26)$ \\
    & \method{} & $\mathbf{166.6}(\mathbf{5.45})$ & $\mathbf{297.5}(\mathbf{5.52})$ & $474.2(\mathbf{5.46})$ & $\mathbf{646.0}(\mathbf{5.50})$ & $\mathbf{635.1}(\mathbf{5.48})$ & $\mathbf{632.3}(\mathbf{5.48})$ \\
    \midrule
    \multirow{2}{*}{\shortstack{HumanEval\\$(T=1)$}} & DFlash & $143.3(4.63)$ & $253.4(4.64)$ & $412.0(4.61)$ & $533.3(4.60)$ & $534.8(4.64)$ & $522.1(4.63)$ \\
    & \method{} & $\mathbf{145.9}(\mathbf{4.79})$ & $\mathbf{257.7}(\mathbf{4.74})$ & $\mathbf{419.7}(\mathbf{4.79})$ & $\mathbf{542.0}(\mathbf{4.75})$ & $\mathbf{536.4}(\mathbf{4.76})$ & $\mathbf{537.0}(\mathbf{4.77})$ \\
    \midrule
    \multirow{2}{*}{\shortstack{LCB\\$(T=0)$}} & DFlash & $168.9(5.37)$ & $288.8(5.41)$ & $449.2(5.45)$ & $500.3(5.44)$ & $482.6(5.45)$ & $487.5(5.45)$ \\
    & \method{} & $\mathbf{185.4}(\mathbf{5.71})$ & $\mathbf{292.4}(\mathbf{5.60})$ & $\mathbf{452.1}(\mathbf{5.61})$ & $\mathbf{501.5}(\mathbf{5.61})$ & $\mathbf{489.6}(\mathbf{5.60})$ & $\mathbf{505.0}(\mathbf{5.65})$ \\
    \midrule
    \multirow{2}{*}{\shortstack{LCB\\$(T=1)$}} & DFlash & $\mathbf{120.4}(4.87)$ & $\mathbf{198.2}(4.92)$ & $304.0(4.89)$ & $339.8(4.87)$ & $330.2(4.89)$ & $323.9(4.87)$ \\
    & \method{} & $120.0(\mathbf{5.00})$ & $194.8(\mathbf{4.97})$ & $\mathbf{306.1}(\mathbf{5.01})$ & $\mathbf{344.7}(\mathbf{5.05})$ & $\mathbf{338.0}(\mathbf{5.00})$ & $\mathbf{325.9}(\mathbf{4.93})$ \\
    \bottomrule
  \end{tabular*}
  \endgroup
\end{table*}

\subsection{Experimental Setup}
\label{sec:exp:setup}

\paragraph{Models and benchmarks.}
We use Qwen3-4B and Qwen3-8B~\citep{yang2025qwen3} as target models, with thinking disabled.
We evaluate math reasoning on GSM8K~\citep{cobbe2021gsm8k}, MATH-500~\citep{hendrycks2021math,lightman2024verify},
and AIME25~\citep{aime25}. We use HumanEval~\citep{chen2021humaneval}, MBPP~\citep{austin2021mbpp},
and LCB~\citep{jain2025livecodebench} for code generation, and MT-Bench~\citep{zheng2023mtbench}
and Alpaca~\citep{taori2023alpaca} for open-ended dialogue.

\paragraph{Drafters.}
On Qwen3-4B, we apply \method{} to four parallel drafters: DFlash~\citep{chen2026dflash},
Domino~\citep{huang2026domino}, DSpark~\citep{dspark2026}, and DFlash2~\citep{dflash2}.
On Qwen3-8B, we apply it to DFlash.
Released checkpoints differ in training data and recipe, so we train each drafter and its
\method{} variant from scratch with the same recipe; within each pair, only the input block
differs.
All drafters have five layers and a block size of $\blk=16$.

\paragraph{Training.}
Our training data include the Qwen3-4B-Instruct-100K dataset provided by the DFlash
authors~\citep{chen2026dflash} and 710K samples from PerfectBlend~\citep{xu2024perfectblend}
with responses regenerated by the target model.
Our drafter implementation uses code from SpecForge~\citep{li2026specforge}.
The target model, including its input embedding and output head, is frozen.
We use AdamW~\citep{loshchilov2019adamw} with a learning rate of $6\times10^{-4}$, cosine decay
with 4\% warmup, gradient clipping at $1.0$, and sequences of up to 3{,}072 tokens.
We set $\gamma=7$ in DFlash's position-decayed loss~\citep{chen2026dflash}.
For drafters trained on Qwen3-4B-Instruct-100K, the global batch is 16 sequences with 448
anchors each, on four NVIDIA RTX 3090 GPUs; for drafters trained on PerfectBlend, it is 14
sequences with 480 anchors each, on eight GPUs.
For each training sequence, we precompute \method{} proposals and their features using the
inference-time retrieval rule.
We train the drafter and zero-initialized copy tables $\mathbf{T}^{\mathrm{len}}$ and
$\mathbf{T}^{\mathrm{agr}}$ together with proposal dropout $p=0.2$.

\paragraph{Inference.}
Unless otherwise specified, all methods run on the Hugging Face Transformers
backend~\citep{wolf2020transformers} with batch size 1, in BF16 with SDPA attention, and generate
at most 2{,}048 new tokens.
The main comparison uses both greedy decoding ($T=0$) and sampling at $T=1$; the remaining
experiments use $T=0$ unless otherwise specified.

\subsection{Main Results}
\label{sec:exp:main}

For Qwen3-4B and Qwen3-8B, we train DFlash and its \method{} counterpart for one epoch on 710K samples
from PerfectBlend.
The four methods are DFlash, pure $n$-gram speculative decoding, DFlash with a switching
baseline (Switch, $L=4$), and DFlash with \method{}.
Pure $n$-gram decoding verifies the retrieved continuation directly.
Switch adapts SAM-Decoding's match-length-based hybrid selection~\citep{hu2024sam} to DFlash.
Pure $n$-gram and Switch retrieve with minimum match length 4.

Table~\ref{tab:exp:main} shows that \method{} achieves the highest average acceptance length
and throughput in all four configurations, reaching 2.93--3.58$\times$ AR throughput.
Appendix~\ref{app:main-results} details the aggregation and uncertainty estimates.
\method{} shows its main advantages on code tasks, with acceptance-length gains over DFlash
of 2.37--3.71\% on HumanEval and 1.43--6.64\% on LCB across the four configurations.
On Qwen3-8B at $T=1$, the LCB gain is accompanied by a 6.58\% increase in throughput.
Compared with Switch, \method{} improves Overall acceptance length by 5.63--6.56\% and
throughput by 1.89--4.63\%.
The direct-copy $n$-gram baseline achieves only 1.19--1.29$\times$ AR throughput overall.

On math benchmarks, \method{} approximately preserves DFlash's acceptance length.
Switch reduces acceptance length by 6.66--8.15\% on GSM8K and 9.64--11.52\% on MATH-500.
These comparisons support letting the drafter learn how to incorporate copy proposals
rather than selecting a draft source solely through a fixed match-length rule.

The gains over the original DFlash remain modest in this 710K setting: 1.08--2.00\% in
Overall acceptance length and 0.39--1.33\% in end-to-end throughput.
Code tasks benefit most, while math and dialogue generally remain closer to the baseline.
This pattern is consistent with $n$-gram proposals relying on useful token repetitions in the context.

\subsection{Generality across Parallel Drafters}
\label{sec:exp:generality}

We train DFlash, Domino, DSpark, and DFlash2, each with and without \method{}, for one epoch
on Qwen3-4B-Instruct-100K.
The target is Qwen3-4B, and evaluation uses $T=0$ on the eight benchmarks.

Table~\ref{tab:exp:generality} shows higher aggregate acceptance length for all four drafters:
the relative gains are 5.08\% for DFlash, 3.40\% for Domino, 2.19\% for DSpark, and 3.03\%
for DFlash2.
The corresponding end-to-end throughput gains over each drafter's original version are
4.04\%, 2.56\%, 0.99\%, and 1.83\%, respectively.
The gains are particularly pronounced on code generation benchmarks.
Across the four drafters, acceptance length increases by 5.67--9.03\% on HumanEval and
4.02--8.76\% on LCB.
Together with the positive aggregate gains for all four drafters, these results indicate
broadly consistent benefits of \method{} across the evaluated drafter architectures.
The smaller gains for DFlash trained on 710K samples than in the 100K study suggest that copied proposals
have less room to improve an already strong drafter.

\subsection{Design Ablations}
\label{sec:exp:ablations}

We use DFlash on Qwen3-4B with the same one-epoch Qwen3-4B-Instruct-100K training setup
as in Section~\ref{sec:exp:generality}, and evaluate on the same benchmarks at $T=0$.
Unless varied, the minimum match length is $m=1$ and proposal dropout is $p=0.2$.

\paragraph{Copy proposals and reliability embeddings.}
Table~\ref{tab:exp:components} separates the copy proposal from the two reliability embeddings.
Replacing mask inputs with copy proposals raises aggregate acceptance length from 3.91
for the original DFlash to 4.06, a 3.79\% gain.
Adding the match-length or agreement embedding raises acceptance length to 4.10 in both cases,
providing a further 0.98\% and 0.86\% improvement over proposal-only inputs, respectively.
The full combination reaches the highest aggregate acceptance length of 4.11,
5.08\% above DFlash and 1.25\% above proposal-only inputs.
These results support combining inexpensive copy proposals with learned reliability features.

\paragraph{Minimum match length.}
We sweep $m\in\{1,2,4\}$ for \method{} and online pure $n$-gram decoding,
where $m$ is $L_{\min}$ in Appendix~\ref{app:retrieval}.
For \method{}, we train a separate model with matching training-time and inference-time $m$,
keeping $p=0.2$.
For DFlash + Switch, we vary the switching threshold $L\in\{1,2,4\}$ while keeping the
retrieval minimum at 1.
Figure~\ref{fig:exp:ablations}\subref{fig:exp:match-sweep} shows that \method{} sustains
an aggregate acceptance length of 4.06--4.11 across the scan, corresponding to
3.80--5.08\% gains over the original DFlash.
It also exceeds the two copying baselines at every tested threshold value.
The highest aggregate acceptance length is obtained at $m=1$, supporting our default
of allowing the drafter to use short-match proposals.
Figure~\ref{fig:exp:ablations}\subref{fig:exp:math-sweep} shows the same comparison on
MATH-500 and GSM8K. Across the tested thresholds, \method{} improves acceptance length
on MATH-500 by 2.48--2.59\% and approximately preserves or modestly improves it on GSM8K.
Even at $L=4$, Switch remains 7.45\% and 5.21\% below the original DFlash on these two
benchmarks, respectively. These results highlight a key advantage of \method{} over Switch:
it lets the drafter learn how to incorporate copy proposals into its predictions, while
Switch relies on a hard match-length rule to choose between copying and drafting.

\paragraph{Proposal dropout.}
At fixed $m=1$, we train models with $p\in\{0,0.2,0.4\}$ and keep the remaining recipe fixed.
Figure~\ref{fig:exp:ablations}\subref{fig:exp:dropout-sweep} reports aggregate acceptance lengths
of 4.06, 4.11, and 4.12, respectively.
Compared with training without proposal dropout, $p=0.2$ and $p=0.4$ improve acceptance
length by 1.33\% and 1.46\%.
The two nonzero rates give similar aggregate performance, supporting proposal dropout
as a useful training choice with stable benefits over the tested range.
We use $p=0.2$ as the default.

\subsection{High-Concurrency Serving}
\label{sec:exp:concurrency}

We compare DFlash and its \method{} variant with Qwen3-8B on SGLang v0.5.11~\citep{zheng2024sglang},
using BF16 and tensor parallelism of 1.
Both drafter checkpoints are trained for one epoch on the 710K-sample PerfectBlend dataset.
We evaluate HumanEval and LCB at $T\in\{0,1\}$, with client concurrency
$c\in\{1,2,4,8,16,32\}$ and four repeated runs per configuration.
Each method pair runs on the same physical GPU, with the execution order alternated across runs.
Table~\ref{tab:exp:concurrency} reports the arithmetic means of TPS and per-request acceptance
length across the four runs.

Across the evaluated configurations, \method{} increases aggregate throughput by 1.73\%
and acceptance length by 3.15\% over DFlash.
HumanEval and LCB show throughput gains of 1.83\% and 1.64\%, respectively,
with corresponding acceptance-length gains of 3.33\% and 2.98\%.
Pooling the configurations with client concurrency 16 and 32 gives a throughput gain of 1.92\%.
These gains first take the geometric mean of paired configuration ratios within each run,
then the arithmetic mean across runs.
The throughput gains from \method{} thus persist under concurrent serving.

\subsection{Mechanism Analysis}
\label{sec:exp:mechanisms}

We analyze how \method{} uses copy proposals with Qwen3-8B at $T=0$.
Appendix~\ref{app:mechanisms} provides the evaluation setup and statistical details.

\paragraph{Removing proposals at inference.}
We compare proposal-on and proposal-off inputs using the same \method{} checkpoint,
with the separately trained original DFlash as a reference.
Table~\ref{tab:exp:no-proposal} shows that enabling proposals increases $\bar{k}$ from
3.90 to 4.12, a paired gain of 0.22 tokens.
This also indicates that \method{} largely preserves DFlash's drafting performance even without
copy proposals (3.90 vs.\ 3.93).
When the proposal supplies the correct token at the proposal-off drafter's first error,
54.07\% of these 7{,}557 opportunities become longer correct prefixes.
Retrieved continuations thus complement the drafter's predictions from masks.

\begin{table}[t]
  \centering
  \caption{With and without copy proposals. The two \method{} rows use the same model.
    $\bar{k}$ is the mean number of consecutive correct draft tokens on the same contexts,
    excluding the anchor.}
  \label{tab:exp:no-proposal}
  \small
  \setlength{\tabcolsep}{3pt}
  \begin{tabular*}{0.95\linewidth}{@{\extracolsep{\fill}}llc@{}}
    \toprule
    Trained drafter & Inference input & $\bar{k}$ \\
    \midrule
    DFlash & Masks & 3.93 \\
    DFlash + \method{} & Masks & 3.90 \\
    DFlash + \method{} & Copy proposals & \textbf{4.12} \\
    \bottomrule
  \end{tabular*}
\end{table}

\begin{figure*}[t]
  \centering
  \captionsetup{skip=4pt}
  \captionsetup[subfigure]{skip=2pt}
  \begin{subfigure}[t]{0.49\textwidth}
    \includegraphics[width=\linewidth]{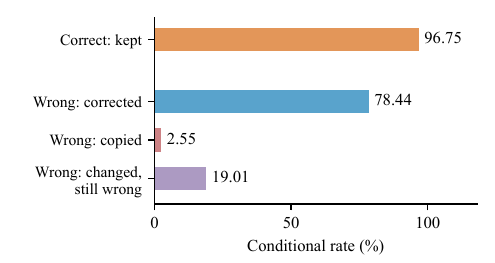}
    \caption{Selective proposal use.}
    \label{fig:exp:mechanism-selectivity}
  \end{subfigure}\hfill
  \begin{subfigure}[t]{0.49\textwidth}
    \includegraphics[width=\linewidth]{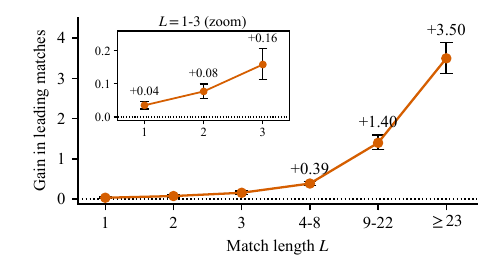}
    \caption{Gain by match length.}
    \label{fig:exp:mechanism-short}
  \end{subfigure}
  \caption{How \method{} uses copy proposals (Qwen3-8B, $T=0$).
    (a) Among correct proposal tokens, the fraction kept; among incorrect ones, the fractions corrected,
    copied, or changed to another wrong token.
    (b) Extra consecutive correct draft tokens with proposals versus without them in the same model,
    grouped by match length $L$. Error bars show 95\% confidence intervals; the inset enlarges $L=1$--$3$.}
  \label{fig:exp:mechanisms}
\end{figure*}

\paragraph{Selective use of proposals.}
Figure~\ref{fig:exp:mechanisms}\subref{fig:exp:mechanism-selectivity} considers proposal
positions whose preceding tokens are correct in both proposal-on and proposal-off drafts.
\method{} retains 96.75\% of correct proposal tokens.
For incorrect proposal tokens, it outputs the correct token in 78.44\% of cases and copies
the incorrect token in only 2.55\%; the remaining 19.01\% are changed to another incorrect token.
Over all proposal positions, including those beyond an earlier draft error, the incorrect-copy
rate is 4.96\%. These behaviors support treating copied tokens as inputs that the drafter can
selectively retain or revise.

\paragraph{Short matches.}
Figure~\ref{fig:exp:mechanisms}\subref{fig:exp:mechanism-short} shows paired gains of
0.04, 0.08, and 0.16 tokens for $L=1,2,3$, respectively.
The gain reaches 3.50 tokens in the $L\ge 23$ group.
This supports allowing short-match proposals and learning how to use them.

\subsection{Overhead of \method{}}
\label{sec:exp:latency}

With Qwen3-8B at $T=0$, \method{} adds a mean of 0.238\,ms per decoding round in
Transformers (Table~\ref{tab:exp:overhead}). Mean per-request decoding time increases by
0.42\%, while end-to-end time increases by 0.43\%.

\begin{table}[t]
  \centering
  \caption{Transformers overhead on identical trajectories. Values are request means with
    95\% bootstrap intervals; total decode uses the ratio of summed times.}
  \label{tab:exp:overhead}
  \small
  \setlength{\tabcolsep}{3pt}
  \begin{tabular}{lcc}
    \toprule
    Metric & Estimate & 95\% CI \\
    \midrule
    Extra decode (ms/round) & 0.238 & $[0.089,\,0.427]$ \\
    Mean decode increase (\%) & 0.42 & $[0.16,\,0.74]$ \\
    End-to-end increase (\%) & 0.43 & $[0.18,\,0.75]$ \\
    Total decode increase (\%) & 0.86 & $[0.30,\,1.67]$ \\
    Index build (ms/request) & 0.355 & $[0.331,\,0.383]$ \\
    \bottomrule
  \end{tabular}
\end{table}

Controlled SGLang measurements at batch sizes 1, 8, and 16 show a similar pattern
(Table~\ref{tab:exp:overhead-sglang}): mean extra latency ranges from 0.223 to 0.557\,ms
per batch round, and paired median wall-time increases are 0.73\%--0.98\%.
These results indicate that \method{} incurs little processing overhead in the tested settings,
including batched decoding. Appendix~\ref{app:latency} gives the measurement setup,
controls, and statistical details.

\begin{table}[t]
  \centering
  \caption{SGLang overhead at fixed batch sizes. Extra time includes amortized index
    initialization (ms/batch-round), with 95\% bootstrap intervals.
    Wall-time increases are paired medians.}
  \label{tab:exp:overhead-sglang}
  \small
  \setlength{\tabcolsep}{3pt}
  \begin{tabular}{rccc}
    \toprule
    Batch & Extra (ms) & 95\% CI & Wall (\%) \\
    \midrule
    1 & 0.223 & $[0.178,\,0.268]$ & 0.73 \\
    8 & 0.557 & $[0.246,\,0.868]$ & 0.98 \\
    16 & 0.534 & $[-0.258,\,1.096]$ & 0.73 \\
    \bottomrule
  \end{tabular}
\end{table}

\section{Conclusion}
\label{sec:conclusion}

\method{} replaces mask inputs to a parallel drafter with continuations retrieved by
$n$-gram matching. In our paired analysis, the drafter kept most correct proposal tokens
and often recovered the correct token when a proposal was wrong, all in its usual
forward pass. Even one-token suffix matches slightly lengthened the
correct prefix. Across eight benchmarks, \method{} increased overall acceptance length and
end-to-end throughput for all four drafters, with the largest gains on code.
Proposal processing added an average of 0.24\,ms per decoding round with Qwen3-8B in Transformers.
Gains were smaller for DFlash trained on 710K samples. Whether other inexpensive proposals
can help a strong drafter when useful repetitions are scarce remains open.

\bibliography{references}

\begin{thebibliography}{73}
\providecommand{\natexlab}[1]{#1}
\providecommand{\url}[1]{\texttt{#1}}
\expandafter\ifx\csname urlstyle\endcsname\relax
  \providecommand{\doi}[1]{doi: #1}\else
  \providecommand{\doi}{doi: \begingroup \urlstyle{rm}\Url}\fi

\bibitem[An et~al.(2026)An, Bai, Liu, Li, and Barsoum]{an2025pard}
An, Z., Bai, H., Liu, Z., Li, D., and Barsoum, E.
\newblock {PARD}: Accelerating {LLM} inference with low-cost {PAR}allel draft
  model adaptation.
\newblock In \emph{International Conference on Learning Representations}, 2026.
\newblock arXiv:2504.18583.

\bibitem[Ankner et~al.(2024)Ankner, Parthasarathy, Nrusimha, Rinard,
  Ragan-Kelley, and Brandon]{ankner2024hydra}
Ankner, Z., Parthasarathy, R., Nrusimha, A., Rinard, C., Ragan-Kelley, J., and
  Brandon, W.
\newblock Hydra: Sequentially-dependent draft heads for {Medusa} decoding.
\newblock In \emph{Conference on Language Modeling}, 2024.

\bibitem[Austin et~al.(2021)Austin, Odena, Nye, Bosma, Michalewski, Dohan,
  Jiang, Cai, Terry, Le, and Sutton]{austin2021mbpp}
Austin, J., Odena, A., Nye, M., Bosma, M., Michalewski, H., Dohan, D., Jiang,
  E., Cai, C., Terry, M., Le, Q., and Sutton, C.
\newblock Program synthesis with large language models.
\newblock \emph{arXiv preprint arXiv:2108.07732}, 2021.

\bibitem[Bhendawade et~al.(2025)Bhendawade, Belousova, Fu, Mason, Lin,
  Rastegari, and Najibi]{bhendawade2025streaming}
Bhendawade, N., Belousova, I., Fu, Q., Mason, H., Lin, A., Rastegari, M., and
  Najibi, M.
\newblock Speculative streaming: Efficient and scalable speculative decoding
  with multi-stream attention.
\newblock In \emph{Conference on Empirical Methods in Natural Language
  Processing}, pp.\  19536--19559, 2025.
\newblock \doi{10.18653/v1/2025.emnlp-main.986}.

\bibitem[Blumer et~al.(1985)Blumer, Blumer, Haussler, Ehrenfeucht, Chen, and
  Seiferas]{blumer1985automaton}
Blumer, A., Blumer, J., Haussler, D., Ehrenfeucht, A., Chen, M.~T., and
  Seiferas, J.
\newblock The smallest automaton recognizing the subwords of a text.
\newblock \emph{Theoretical Computer Science}, 40:\penalty0 31--55, 1985.
\newblock \doi{10.1016/0304-3975(85)90157-4}.

\bibitem[Cai et~al.(2024)Cai, Li, Geng, Peng, Lee, Chen, and
  Dao]{cai2024medusa}
Cai, T., Li, Y., Geng, Z., Peng, H., Lee, J.~D., Chen, D., and Dao, T.
\newblock Medusa: Simple {LLM} inference acceleration framework with multiple
  decoding heads.
\newblock In \emph{International Conference on Machine Learning}, pp.\
  5209--5235, 2024.
\newblock URL \url{https://proceedings.mlr.press/v235/cai24b.html}.

\bibitem[Chen et~al.(2023)Chen, Borgeaud, Irving, Lespiau, Sifre, and
  Jumper]{chen2023accelerating}
Chen, C., Borgeaud, S., Irving, G., Lespiau, J.-B., Sifre, L., and Jumper, J.
\newblock Accelerating large language model decoding with speculative sampling.
\newblock \emph{arXiv preprint arXiv:2302.01318}, 2023.

\bibitem[Chen et~al.(2026)Chen, Liang, and Liu]{chen2026dflash}
Chen, J., Liang, Y., and Liu, Z.
\newblock {DFlash}: Block diffusion for flash speculative decoding.
\newblock In \emph{International Conference on Machine Learning}, 2026.

\bibitem[Chen et~al.(2021)]{chen2021humaneval}
Chen, M. et~al.
\newblock Evaluating large language models trained on code.
\newblock \emph{arXiv preprint arXiv:2107.03374}, 2021.

\bibitem[Cheng et~al.(2026)Cheng, Yu, Shao, Li, Xiong, Qian, Zhu, Ma, Zhang,
  Ye, Chen, Deng, Yu, Dai, Zhang, Wei, Tan, Yang, Xu, Wu, Xu, Wang, Chen, Tian,
  Bi, Hao, Chen, Cao, Zhang, Xu, Zhang, Zhao, and Liang]{dspark2026}
Cheng, X., Yu, X., Shao, C., Li, J., Xiong, Y., Qian, Y., Zhu, J., Ma, S.,
  Zhang, X., Ye, J., Chen, Q., Deng, C., Yu, J., Dai, D., Zhang, Z., Wei, Y.,
  Tan, Y., Yang, W., Xu, R., Wu, Y., Xu, Z., Wang, X., Chen, M., Tian, R., Bi,
  X., Hao, Z., Chen, S., Cao, H., Zhang, W., Xu, A., Zhang, H., Zhao, D., and
  Liang, W.
\newblock {DSpark}: Confidence-scheduled speculative decoding with
  semi-autoregressive generation.
\newblock \emph{arXiv preprint arXiv:2607.05147}, 2026.

\bibitem[Cheng et~al.(2024)Cheng, Zhang, Zhang, Wang, and
  Wang]{cheng2024redrafter}
Cheng, Y., Zhang, A., Zhang, X., Wang, C., and Wang, Y.
\newblock Recurrent drafter for fast speculative decoding in large language
  models.
\newblock \emph{arXiv preprint arXiv:2403.09919}, 2024.

\bibitem[Cobbe et~al.(2021)Cobbe, Kosaraju, Bavarian, Chen, Jun, Kaiser,
  Plappert, Tworek, Hilton, Nakano, Hesse, and Schulman]{cobbe2021gsm8k}
Cobbe, K., Kosaraju, V., Bavarian, M., Chen, M., Jun, H., Kaiser, L., Plappert,
  M., Tworek, J., Hilton, J., Nakano, R., Hesse, C., and Schulman, J.
\newblock Training verifiers to solve math word problems.
\newblock \emph{arXiv preprint arXiv:2110.14168}, 2021.

\bibitem[{DeepSeek-AI}(2024)]{deepseek2024v3}
{DeepSeek-AI}.
\newblock {DeepSeek-V3} technical report.
\newblock \emph{arXiv preprint arXiv:2412.19437}, 2024.

\bibitem[Eisenstadt et~al.(2026)Eisenstadt, Lidar, Gonen, Luxembourg, Zimerman,
  Wolf, David, and Naiman]{eisenstadt2026dcoco}
Eisenstadt, R., Lidar, H., Gonen, T., Luxembourg, O., Zimerman, I., Wolf, L.,
  David, E., and Naiman, I.
\newblock Contextual correction of parallel drafts for speculative decoding.
\newblock \emph{arXiv preprint arXiv:2610.12570}, 2026.

\bibitem[Elhoushi et~al.(2024)Elhoushi, Shrivastava, Liskovich, Hosmer, Wasti,
  Lai, Mahmoud, Acun, Agarwal, Roman, Aly, Chen, and Wu]{elhoushi2024layerskip}
Elhoushi, M., Shrivastava, A., Liskovich, D., Hosmer, B., Wasti, B., Lai, L.,
  Mahmoud, A., Acun, B., Agarwal, S., Roman, A., Aly, A., Chen, B., and Wu,
  C.-J.
\newblock {LayerSkip}: Enabling early exit inference and self-speculative
  decoding.
\newblock In \emph{Annual Meeting of the Association for Computational
  Linguistics}, 2024.
\newblock \doi{10.18653/v1/2024.acl-long.681}.

\bibitem[Field \& Welsh(2007)Field and Welsh]{field2007cluster}
Field, C.~A. and Welsh, A.~H.
\newblock Bootstrapping clustered data.
\newblock \emph{Journal of the Royal Statistical Society: Series B (Statistical
  Methodology)}, 69\penalty0 (3):\penalty0 369--390, 2007.
\newblock \doi{10.1111/j.1467-9868.2007.00593.x}.

\bibitem[Fu et~al.(2024)Fu, Bailis, Stoica, and Zhang]{fu2024lookahead}
Fu, Y., Bailis, P., Stoica, I., and Zhang, H.
\newblock Break the sequential dependency of {LLM} inference using lookahead
  decoding.
\newblock In \emph{International Conference on Machine Learning}, 2024.

\bibitem[Gao et~al.(2025)Gao, Xie, Xiang, and Ji]{gao2025falcon}
Gao, X., Xie, W., Xiang, Y., and Ji, F.
\newblock Falcon: Faster and parallel inference of large language models
  through enhanced semi-autoregressive drafting and custom-designed decoding
  tree.
\newblock In \emph{AAAI Conference on Artificial Intelligence}, 2025.
\newblock \doi{10.1609/aaai.v39i22.34566}.

\bibitem[Ghazvininejad et~al.(2019)Ghazvininejad, Levy, Liu, and
  Zettlemoyer]{ghazvininejad2019maskpredict}
Ghazvininejad, M., Levy, O., Liu, Y., and Zettlemoyer, L.
\newblock Mask-predict: Parallel decoding of conditional masked language
  models.
\newblock In \emph{Conference on Empirical Methods in Natural Language
  Processing}, 2019.

\bibitem[Girshick et~al.(2014)Girshick, Donahue, Darrell, and
  Malik]{girshick2014rcnn}
Girshick, R., Donahue, J., Darrell, T., and Malik, J.
\newblock Rich feature hierarchies for accurate object detection and semantic
  segmentation.
\newblock In \emph{IEEE Conference on Computer Vision and Pattern Recognition},
  2014.

\bibitem[Gu et~al.(2016)Gu, Lu, Li, and Li]{gu2016copynet}
Gu, J., Lu, Z., Li, H., and Li, V. O.~K.
\newblock Incorporating copying mechanism in sequence-to-sequence learning.
\newblock In \emph{Annual Meeting of the Association for Computational
  Linguistics}, 2016.

\bibitem[Guu et~al.(2018)Guu, Hashimoto, Oren, and Liang]{guu2018editing}
Guu, K., Hashimoto, T.~B., Oren, Y., and Liang, P.
\newblock Generating sentences by editing prototypes.
\newblock \emph{Transactions of the Association for Computational Linguistics},
  2018.

\bibitem[Hashimoto et~al.(2018)Hashimoto, Guu, Oren, and
  Liang]{hashimoto2018retrieve}
Hashimoto, T.~B., Guu, K., Oren, Y., and Liang, P.
\newblock A retrieve-and-edit framework for predicting structured outputs.
\newblock In \emph{Advances in Neural Information Processing Systems}, 2018.

\bibitem[He et~al.(2024)He, Zhong, Cai, Lee, and He]{he2024rest}
He, Z., Zhong, Z., Cai, T., Lee, J.~D., and He, D.
\newblock {REST}: Retrieval-based speculative decoding.
\newblock In \emph{North American Chapter of the Association for Computational
  Linguistics}, 2024.

\bibitem[Hendrycks et~al.(2021)Hendrycks, Burns, Kadavath, Arora, Basart, Tang,
  Song, and Steinhardt]{hendrycks2021math}
Hendrycks, D., Burns, C., Kadavath, S., Arora, A., Basart, S., Tang, E., Song,
  D., and Steinhardt, J.
\newblock Measuring mathematical problem solving with the {MATH} dataset.
\newblock In \emph{NeurIPS Datasets and Benchmarks Track}, 2021.

\bibitem[Hu et~al.(2025)Hu, Wang, Zhang, Zhang, Li, Chen, and Zhang]{hu2024sam}
Hu, Y., Wang, K., Zhang, X., Zhang, F., Li, C., Chen, H., and Zhang, J.
\newblock {SAM} decoding: Speculative decoding via suffix automaton.
\newblock In \emph{Proceedings of the 63rd Annual Meeting of the Association
  for Computational Linguistics (Volume 1: Long Papers)}, pp.\  12187--12204,
  2025.
\newblock \doi{10.18653/v1/2025.acl-long.595}.
\newblock URL \url{https://aclanthology.org/2025.acl-long.595/}.

\bibitem[Huang et~al.(2026)Huang, Zhang, Zhang, Lin, Xu, and
  Zhang]{huang2026domino}
Huang, J., Zhang, Y., Zhang, Q., Lin, H., Xu, H., and Zhang, L.
\newblock Domino: Decoupling causal modeling from autoregressive drafting in
  speculative decoding.
\newblock \emph{arXiv preprint arXiv:2605.29707}, 2026.

\bibitem[Hui et~al.(2026)Hui, Huang, Campos~Salas, Sun, Pemberton, Song,
  Khetan, and Karypis]{hui2026peagle}
Hui, M., Huang, X., Campos~Salas, J., Sun, Y., Pemberton, N., Song, X., Khetan,
  A., and Karypis, G.
\newblock {P-EAGLE}: Parallel-drafting {EAGLE} with scalable training.
\newblock \emph{arXiv preprint arXiv:2602.01469}, 2026.

\bibitem[{Inco AI}(2026)]{dflash2}
{Inco AI}.
\newblock {DFlash 2: Keep Drafting Parallel}, August 2026.
\newblock URL \url{https://inco.ai/blog/dflash2/}.

\bibitem[Jain et~al.(2025)Jain, Han, Gu, Li, Yan, Zhang, Wang, Solar-Lezama,
  Sen, and Stoica]{jain2025livecodebench}
Jain, N., Han, K., Gu, A., Li, W.-D., Yan, F., Zhang, T., Wang, S.,
  Solar-Lezama, A., Sen, K., and Stoica, I.
\newblock {LiveCodeBench}: Holistic and contamination free evaluation of large
  language models for code.
\newblock In \emph{International Conference on Learning Representations}, 2025.

\bibitem[Kim et~al.(2024)Kim, Suresh, Papineni, Riley, Kumar, and
  Benton]{kim2024refinement}
Kim, T., Suresh, A.~T., Papineni, K., Riley, M., Kumar, S., and Benton, A.
\newblock Accelerating blockwise parallel language models with draft
  refinement.
\newblock In \emph{Advances in Neural Information Processing Systems},
  volume~37, 2024.
\newblock \doi{10.52202/079017-1081}.

\bibitem[Leviathan et~al.(2023)Leviathan, Kalman, and
  Matias]{leviathan2023fast}
Leviathan, Y., Kalman, M., and Matias, Y.
\newblock Fast inference from transformers via speculative decoding.
\newblock In \emph{International Conference on Machine Learning}, 2023.

\bibitem[Li et~al.(2026{\natexlab{a}})Li, Fu, Fang, Zhao, Tang, Yuan, and
  Wang]{li2026diffuspec}
Li, G., Fu, Z., Fang, M., Zhao, Q., Tang, M., Yuan, C., and Wang, J.
\newblock {DiffuSpec}: Unlocking diffusion language models for speculative
  decoding.
\newblock In \emph{Findings of ACL 2026}, pp.\  20896--20910,
  2026{\natexlab{a}}.
\newblock \doi{10.18653/v1/2026.findings-acl.1048}.

\bibitem[Li et~al.(2026{\natexlab{b}})Li, Wang, Zhu, Wang, Yin, Shi, Chen,
  Dong, Chen, Pan, Li, Xie, Zhang, Yu, Wen, Tsang, and Zhang]{li2026specforge}
Li, S., Wang, C., Zhu, Y., Wang, Y., Yin, F., Shi, S., Chen, Y., Dong, X.,
  Chen, Q., Pan, J., Li, J., Xie, L., Zhang, Y., Yu, L., Wen, Y., Tsang, I.,
  and Zhang, T.
\newblock {SpecForge}: A flexible and efficient open-source training framework
  for speculative decoding.
\newblock \emph{arXiv preprint arXiv:2603.18567}, 2026{\natexlab{b}}.

\bibitem[Li et~al.(2024{\natexlab{a}})Li, Wei, Zhang, and Zhang]{li2024eagle}
Li, Y., Wei, F., Zhang, C., and Zhang, H.
\newblock {EAGLE}: Speculative sampling requires rethinking feature
  uncertainty.
\newblock In \emph{International Conference on Machine Learning},
  2024{\natexlab{a}}.

\bibitem[Li et~al.(2024{\natexlab{b}})Li, Wei, Zhang, and Zhang]{li2024eagle2}
Li, Y., Wei, F., Zhang, C., and Zhang, H.
\newblock {EAGLE-2}: Faster inference of language models with dynamic draft
  trees.
\newblock In \emph{Conference on Empirical Methods in Natural Language
  Processing}, 2024{\natexlab{b}}.

\bibitem[Li et~al.(2025)Li, Wei, Zhang, and Zhang]{li2025eagle3}
Li, Y., Wei, F., Zhang, C., and Zhang, H.
\newblock {EAGLE-3}: Scaling up inference acceleration of large language models
  via training-time test.
\newblock \emph{arXiv preprint arXiv:2503.01840}, 2025.

\bibitem[Lightman et~al.(2024)Lightman, Kosaraju, Burda, Edwards, Baker, Lee,
  Leike, Schulman, Sutskever, and Cobbe]{lightman2024verify}
Lightman, H., Kosaraju, V., Burda, Y., Edwards, H., Baker, B., Lee, T., Leike,
  J., Schulman, J., Sutskever, I., and Cobbe, K.
\newblock Let's verify step by step.
\newblock In \emph{International Conference on Learning Representations}, 2024.

\bibitem[Lin et~al.(2024)]{lin2024bita}
Lin, F. et~al.
\newblock {BiTA}: Bi-directional tuning for lossless acceleration in large
  language models.
\newblock \emph{arXiv preprint arXiv:2401.12522}, 2024.

\bibitem[Lin et~al.(2026)Lin, Zeng, Peng, Liu, and Ji]{lin2026retrace}
Lin, L., Zeng, Z., Peng, S., Liu, S., and Ji, R.
\newblock {ReTrace}: Rejected-trajectory conditioning for speculative decoding.
\newblock \emph{arXiv preprint arXiv:2608.29748}, 2026.

\bibitem[Liu et~al.(2026)Liu, Li, Zhao, Gao, Zhou, Zhang, Dou, Zhong, Wang, and
  Tian]{liu2026dart}
Liu, F., Li, X., Zhao, K., Gao, Y., Zhou, Z., Zhang, Z., Dou, W., Zhong, S.,
  Wang, Z., and Tian, C.
\newblock {DART}: Low-latency parallel drafting with continuity-aware tree
  pruning for speculative decoding.
\newblock In \emph{Proceedings of the 2026 Conference on Empirical Methods in
  Natural Language Processing}, pp.\  11316--11331, 2026.
\newblock \doi{10.18653/2026.emnlp-main.492}.

\bibitem[Loshchilov \& Hutter(2019)Loshchilov and Hutter]{loshchilov2019adamw}
Loshchilov, I. and Hutter, F.
\newblock Decoupled weight decay regularization.
\newblock In \emph{International Conference on Learning Representations}, 2019.

\bibitem[Luo et~al.(2025)]{luo2025tokenrecycling}
Luo, X. et~al.
\newblock Turning trash into treasure: Accelerating inference of large language
  models with token recycling.
\newblock In \emph{Annual Meeting of the Association for Computational
  Linguistics}, 2025.

\bibitem[Monea et~al.(2023)Monea, Joulin, and Grave]{monea2023pass}
Monea, G., Joulin, A., and Grave, E.
\newblock {PaSS}: Parallel speculative sampling.
\newblock \emph{arXiv preprint arXiv:2311.13581}, 2023.

\bibitem[Oliaro et~al.(2025)Oliaro, Jia, Campos, and
  Qiao]{oliaro2025suffixdecoding}
Oliaro, G., Jia, Z., Campos, D., and Qiao, A.
\newblock {SuffixDecoding}: Extreme speculative decoding for emerging {AI}
  applications.
\newblock In \emph{Advances in Neural Information Processing Systems}, 2025.

\bibitem[Quan et~al.(2025)Quan, Feng, Hao, Jiang, Zhang, and
  Wang]{quan2025rasd}
Quan, G., Feng, W., Hao, C., Jiang, G., Zhang, Y., and Wang, H.~H.
\newblock {RASD}: Retrieval-augmented speculative decoding.
\newblock In \emph{Findings of the Association for Computational Linguistics:
  ACL 2025}, pp.\  6167--6177, 2025.
\newblock \doi{10.18653/v1/2025.findings-acl.320}.

\bibitem[Ranjan \& Venkitesh(2026)Ranjan and Venkitesh]{ranjan2026hybridspark}
Ranjan, E. and Venkitesh, B.
\newblock {Hybrid Spark}: Adaptive speculative decoding that lets the content
  choose the drafter.
\newblock \emph{arXiv preprint arXiv:2611.05611}, 2026.

\bibitem[Ren et~al.(2015)Ren, He, Girshick, and Sun]{ren2015faster}
Ren, S., He, K., Girshick, R., and Sun, J.
\newblock Faster {R-CNN}: Towards real-time object detection with region
  proposal networks.
\newblock In \emph{Advances in Neural Information Processing Systems}, 2015.

\bibitem[Ringel \& Romano(2026)Ringel and Romano]{ringel2026ddtree}
Ringel, L. and Romano, Y.
\newblock Accelerating speculative decoding with block diffusion draft trees.
\newblock \emph{arXiv preprint arXiv:2604.12989}, 2026.

\bibitem[Saxena(2023)]{saxena2023pld}
Saxena, A.
\newblock Prompt lookup decoding.
\newblock \url{https://github.com/apoorvumang/prompt-lookup-decoding}, 2023.

\bibitem[Shen \& Huang(2026)Shen and Huang]{shen2026dmtp}
Shen, Y. and Huang, J.
\newblock {DMTP}: A better first token for block drafting via {MTP}.
\newblock \url{https://sylvan820.github.io/D-MTP/}, 2026.
\newblock Technical blog.

\bibitem[Shen et~al.(2026{\natexlab{a}})Shen, Liu, Hu, Kong, Zhang, Dai, Zhang,
  Ge, Chen, Li, Wan, and Wang]{graft2026}
Shen, Y., Liu, T., Hu, X., Kong, Q., Zhang, B., Dai, J., Zhang, J., Ge, S.,
  Chen, L., Li, Y., Wan, M., and Wang, C.
\newblock Draft less, retrieve more: Hybrid tree construction for speculative
  decoding.
\newblock \emph{arXiv preprint arXiv:2605.20104}, 2026{\natexlab{a}}.

\bibitem[Shen et~al.(2026{\natexlab{b}})Shen, Liu, Shen, Wu, Kong, Li, and
  Wang]{shen2026double}
Shen, Y., Liu, T., Shen, J., Wu, J., Kong, Q., Li, H., and Wang, C.
\newblock {Double}: Breaking the acceleration limit via double retrieval
  speculative parallelism.
\newblock In \emph{Proceedings of the 64th Annual Meeting of the Association
  for Computational Linguistics (Volume 1: Long Papers)}, pp.\  19242--19263,
  2026{\natexlab{b}}.
\newblock \doi{10.18653/v1/2026.acl-long.879}.
\newblock URL \url{https://aclanthology.org/2026.acl-long.879/}.

\bibitem[Stern et~al.(2018)Stern, Shazeer, and Uszkoreit]{stern2018blockwise}
Stern, M., Shazeer, N., and Uszkoreit, J.
\newblock Blockwise parallel decoding for deep autoregressive models.
\newblock In \emph{Advances in Neural Information Processing Systems},
  volume~31, 2018.

\bibitem[Su et~al.(2026)Su, Majchrowski, Yu, Shah, Rogawski, Morkisz,
  Bhiwandiwalla, and Howard]{su2026hybridverified}
Su, X., Majchrowski, D., Yu, F., Shah, V.~A., Rogawski, S., Morkisz, P.,
  Bhiwandiwalla, A., and Howard, P.
\newblock {Hybrid Verified Decoding}: Learning to allocate verification in
  speculative decoding.
\newblock \emph{arXiv preprint arXiv:2606.01019}, 2026.

\bibitem[Taori et~al.(2023)Taori, Gulrajani, Zhang, Dubois, Li, Guestrin,
  Liang, and Hashimoto]{taori2023alpaca}
Taori, R., Gulrajani, I., Zhang, T., Dubois, Y., Li, X., Guestrin, C., Liang,
  P., and Hashimoto, T.~B.
\newblock Stanford alpaca: An instruction-following {LLaMA} model.
\newblock \url{https://github.com/tatsu-lab/stanford_alpaca}, 2023.

\bibitem[Vinyals et~al.(2015)Vinyals, Fortunato, and
  Jaitly]{vinyals2015pointer}
Vinyals, O., Fortunato, M., and Jaitly, N.
\newblock Pointer networks.
\newblock In \emph{Advances in Neural Information Processing Systems}, 2015.

\bibitem[Wang et~al.(2026)Wang, Wertheimer, Lim, Srivatsa, Ganti, Zhang, and
  Wang]{xpress2026}
Wang, Z., Wertheimer, D., Lim, Y. C.~F., Srivatsa, M., Ganti, R.~K., Zhang, M.,
  and Wang, N.
\newblock {XPress}: Parallel refinement for diffusion drafters in speculative
  decoding.
\newblock \emph{arXiv preprint arXiv:2608.02438}, 2026.

\bibitem[Wolf et~al.(2020)]{wolf2020transformers}
Wolf, T. et~al.
\newblock Transformers: State-of-the-art natural language processing.
\newblock In \emph{Proceedings of the 2020 Conference on Empirical Methods in
  Natural Language Processing: System Demonstrations}, 2020.

\bibitem[Xiao et~al.(2024)Xiao, Zhang, Ge, Ouyang, Ordonez, and
  Yu]{xiao2024parallelspec}
Xiao, Z., Zhang, H., Ge, T., Ouyang, S., Ordonez, V., and Yu, D.
\newblock {ParallelSpec}: Parallel drafter for efficient speculative decoding.
\newblock \emph{arXiv preprint arXiv:2410.05589}, 2024.

\bibitem[Xu et~al.(2023)Xu, Crego, and Yvon]{xu2023tmlevt}
Xu, J., Crego, J., and Yvon, F.
\newblock Integrating translation memories into non-autoregressive machine
  translation.
\newblock In \emph{Proceedings of the 17th Conference of the European Chapter
  of the Association for Computational Linguistics}, pp.\  1326--1338, 2023.
\newblock \doi{10.18653/v1/2023.eacl-main.96}.

\bibitem[Xu et~al.(2024)Xu, Helenowski, Sankararaman, Jin, Peng, Han, Nie, Zhu,
  Zhang, Zhou, et~al.]{xu2024perfectblend}
Xu, T., Helenowski, E., Sankararaman, K.~A., Jin, D., Peng, K., Han, E., Nie,
  S., Zhu, C., Zhang, H., Zhou, W., et~al.
\newblock The perfect blend: Redefining {RLHF} with mixture of judges.
\newblock \emph{arXiv preprint arXiv:2409.20370}, 2024.

\bibitem[Yang et~al.(2025)]{yang2025qwen3}
Yang, A. et~al.
\newblock Qwen3 technical report.
\newblock \emph{arXiv preprint arXiv:2505.09388}, 2025.

\bibitem[Yang et~al.(2023)Yang, Ge, Wang, Jiao, Jiang, Yang, Majumder, and
  Wei]{yang2023llma}
Yang, N., Ge, T., Wang, L., Jiao, B., Jiang, D., Yang, L., Majumder, R., and
  Wei, F.
\newblock Inference with reference: Lossless acceleration of large language
  models.
\newblock \emph{arXiv preprint arXiv:2304.04487}, 2023.

\bibitem[Yu et~al.(2026)]{yu2026dedit}
Yu, L. et~al.
\newblock {DEdit}: Iterative draft editing for speculative decoding.
\newblock \emph{arXiv preprint arXiv:2609.38510}, 2026.

\bibitem[Zhang et~al.(2026)Zhang, Chen, Sun, Gu, Qiao, Sun, and
  Jiang]{zhang2026dspine}
Zhang, H., Chen, K., Sun, H., Gu, W., Qiao, R., Sun, X., and Jiang, B.
\newblock Draft in parallel, condition through depth: Adjacent causal injection
  for speculative decoding.
\newblock \emph{arXiv preprint arXiv:2609.36173}, 2026.

\bibitem[Zhang et~al.(2024)Zhang, Wang, Li, Shou, Chen, Chen, and
  Mehrotra]{zhang2024draftverify}
Zhang, J., Wang, J., Li, H., Shou, L., Chen, K., Chen, G., and Mehrotra, S.
\newblock {Draft \& Verify}: Lossless large language model acceleration via
  self-speculative decoding.
\newblock In \emph{Annual Meeting of the Association for Computational
  Linguistics}, 2024.
\newblock \doi{10.18653/v1/2024.acl-long.607}.

\bibitem[Zhang \& {Math-AI Team}(2025)Zhang and {Math-AI Team}]{aime25}
Zhang, Y. and {Math-AI Team}.
\newblock American invitational mathematics examination ({AIME}) 2025.
\newblock \url{https://huggingface.co/datasets/math-ai/aime25}, 2025.

\bibitem[Zhao et~al.(2024)Zhao, Huang, Han, Xu, Xiao, Zhang, Fang, Zhang, Liu,
  and Sun]{zhao2024ouroboros}
Zhao, W., Huang, Y., Han, X., Xu, W., Xiao, C., Zhang, X., Fang, Y., Zhang, K.,
  Liu, Z., and Sun, M.
\newblock {Ouroboros}: Generating longer drafts phrase by phrase for faster
  speculative decoding.
\newblock In \emph{Proceedings of the 2024 Conference on Empirical Methods in
  Natural Language Processing}, pp.\  13378--13393, 2024.
\newblock \doi{10.18653/v1/2024.emnlp-main.742}.
\newblock URL \url{https://aclanthology.org/2024.emnlp-main.742/}.

\bibitem[Zheng \& Li(2026)Zheng and Li]{zheng2026dels}
Zheng, H.-K. and Li, P.
\newblock {DeLS-Spec}: Decoupled long-short contexts for parallel speculative
  drafting.
\newblock \emph{arXiv preprint arXiv:2607.07409}, 2026.

\bibitem[Zheng et~al.(2023)Zheng, Chiang, Sheng, Zhuang, Wu, Zhuang, Lin, Li,
  Li, Xing, Zhang, Gonzalez, and Stoica]{zheng2023mtbench}
Zheng, L., Chiang, W.-L., Sheng, Y., Zhuang, S., Wu, Z., Zhuang, Y., Lin, Z.,
  Li, Z., Li, D., Xing, E.~P., Zhang, H., Gonzalez, J.~E., and Stoica, I.
\newblock Judging {LLM}-as-a-judge with {MT-Bench} and chatbot arena.
\newblock In \emph{NeurIPS Datasets and Benchmarks Track}, 2023.

\bibitem[Zheng et~al.(2024)Zheng, Yin, Xie, Sun, Huang, Yu, Cao, Kozyrakis,
  Stoica, Gonzalez, Barrett, and Sheng]{zheng2024sglang}
Zheng, L., Yin, L., Xie, Z., Sun, C., Huang, J., Yu, C.~H., Cao, S., Kozyrakis,
  C., Stoica, I., Gonzalez, J.~E., Barrett, C., and Sheng, Y.
\newblock {SGLang}: Efficient execution of structured language model programs.
\newblock In \emph{Advances in Neural Information Processing Systems},
  volume~37, 2024.

\bibitem[Zhou et~al.(2024)Zhou, Lyu, Rawat, Menon, Rostamizadeh, Kumar, Kagy,
  and Agarwal]{zhou2024distillspec}
Zhou, Y., Lyu, K., Rawat, A.~S., Menon, A.~K., Rostamizadeh, A., Kumar, S.,
  Kagy, J.-F., and Agarwal, R.
\newblock {DistillSpec}: Improving speculative decoding via knowledge
  distillation.
\newblock In \emph{International Conference on Learning Representations}, 2024.

\end{thebibliography}
\bibliographystyle{icml2026}

\startpaperappendix
\section{Copy proposal retrieval}
\label{app:retrieval}

We specify the retrieval rule from Section~\ref{sec:method:retrieval} for the committed
sequence $x_{1:a}$.

Let $\mathcal{C}$ be the set of control tokens, i.e.\ the tokenizer's special tokens such as
role markers, end-of-turn and reasoning delimiters.
Matches and proposals never contain a control token, so that a proposal cannot run across a turn
boundary.
For a suffix length $\ell$, write $S_\ell = x_{a-\ell+1:a}$.
An earlier occurrence of $S_\ell$ is an end position $j < a$ with $x_{j-\ell+1:j} = S_\ell$.
Let $d(j)$ be the number of tokens strictly between position $j$ and the next control
token in the committed sequence, or $+\infty$ if no control token follows $j$.
The \emph{available continuation length} at $j$ is then
\begin{equation}
  w(j) = \min\{\blk-1,\; a-j,\; d(j)\},
\end{equation}
so a continuation never extends beyond the committed sequence ($j + w(j) \le a$) and never
crosses a control token.
The occurrence may overlap the suffix itself.
The eligible pairs of suffix length and earlier end position form the set
\[
  \mathcal{A} =
  \left\{(\ell,j)\;\middle|\;
  \begin{aligned}
    &L_{\min} \le \ell \le j < a,\\
    &x_{j-\ell+1:j} = S_\ell,\quad w(j) \ge 1,\\
    &\{x_{a-\ell+1},\ldots,x_a\} \cap \mathcal{C} = \varnothing
  \end{aligned}
  \right\}.
\]
If $\mathcal{A} \ne \varnothing$, we select
\begin{equation}
  (L,j^\star) =
  \operatorname*{arg\,max}^{\mathrm{lex}}_{(\ell,j)\in\mathcal{A}}
  \bigl(\ell, w(j), j\bigr),
  \label{eq:retrieval-rule}
\end{equation}
where $\mathrm{lex}$ denotes lexicographic order.
The proposal in Eq.~\eqref{eq:retrieval-proposal} has length $\nslot = w(j^\star)$.
If $\mathcal{A} = \varnothing$, $\nslot=0$ and the drafter receives the unmodified
block~\eqref{eq:plain-block}.
We use $L_{\min}=1$ and leave it to the learned features (\S\ref{sec:method:features}) to tell
the drafter how much a short match is worth.

\paragraph{Online index.}
Retrieval runs over the prompt and the generated text and must keep up with decoding.
Following \citet{hu2024sam}, we index the committed sequence with an online suffix
automaton~\citep{blumer1985automaton}, inserting each newly committed token once.
Besides the usual transitions and suffix links, each automaton state $s$ stores two kinds of
metadata.
The first is, for every proposal length $w \in \{1, \dots, \blk-1\}$, the most recent end position
of an occurrence of $s$ with at least $w$ available continuation tokens.
A record for length $w$ is written as soon as the $w$-th continuation token is committed and is
propagated along suffix links until it meets a record that is already at least as recent.
The second is the number of end positions of $s$, which is incremented along the suffix-link chain
of each appended token and is used by the agreement feature below.
A query walks from the state of the longest suffix free of control tokens towards shorter suffixes and stops
at the first state that holds a record, scanning $w$ from $\blk-1$ down to $1$.
This returns
exactly the proposal defined by Eq.~\eqref{eq:retrieval-rule}.

For the agreement feature in Section~\ref{sec:method:features}, $n$ is the end-position count
of the matched state minus one to exclude the current suffix. The count $n_k$ is the
end-position count of the state reached by following the transitions for $c_{1:k}$ from
the matched state.

\section{Reliability feature buckets}
\label{app:feature-buckets}

\paragraph{Match length.}
The length bucket $b(L)$ takes $N_{\mathrm{len}}=21$ values. Lengths
$L=1,\dots,8$ each have their own bucket. The buckets then become progressively
wider, and the final bucket covers $L \ge 45$.

\paragraph{Agreement.}
Of the five agreement buckets, bucket $0$ is reserved for $n=1$. When $n>1$,
the agreement score maps to a bucket by
\begin{equation}
  \operatorname{Bucket}_{\mathrm{agr}}(\alpha_k) =
  \begin{cases}
    1, & \alpha_k = 0 \quad \text{(no other occurrence agrees)},\\
    2, & 0 < \alpha_k < \tfrac12,\\
    3, & \tfrac12 \le \alpha_k < 1,\\
    4, & \alpha_k = 1 \quad \text{(all other occurrences agree)}.
  \end{cases}
\end{equation}

\paragraph{Parameter count.}
With $\blk=16$, the two tables add $15 \times (21 + 5) \times d$ parameters,
about $1.0$M for $d=2560$ and $1.6$M for $d=4096$.

\section{Additional mechanism analyses}
\label{app:mechanisms}

\subsection{Evaluation setup}

We use frozen DFlash and \method{} checkpoints with Qwen3-8B, trained on PerfectBlend 710K.
We run the replay in Transformers at $T=0$ with block size 16.
The reference set contains 994 requests from 914 conversations across the eight benchmarks.
The replay covers 150{,}140 distinct positions, including archived DFlash round starts and
uniformly sampled positions. Our primary subset contains 93{,}733 archived round starts
from 907 conversations, with a complete 15-token reference continuation that agrees with
the teacher-forced target argmax at every position.
The target features and reference prefix are fixed across each intervention pair.
The proposal-off condition replaces proposal inputs with masks using the same \method{} weights.
For predictions $\hat y_{1:15}$ and reference tokens $y_{1:15}$, the diagnostic is
\begin{equation}
  k = \max\{j\in\{0,\ldots,15\}:\hat y_{1:j}=y_{1:j}\}.
\end{equation}
This count excludes the anchor and is distinct from online acceptance length $\tau$.
All paired comparisons use fresh predictions at the same recorded contexts.
Reported means pool anchors. For 95\% intervals we use a cluster bootstrap~\citep{field2007cluster}
with 2{,}000 resamples, keeping all anchors from each sampled conversation together.

\subsection{Proposal behavior and match length}

The conditional rates in Figure~\ref{fig:exp:mechanisms}\subref{fig:exp:mechanism-selectivity}
include a supplied proposal position only if all preceding draft tokens match the reference
in both proposal-on and proposal-off predictions; the position itself may be incorrect.
There are 107{,}535 correct and 199{,}403 incorrect proposal tokens in this subset.
The model retains 104{,}039 of the correct tokens, corrects 156{,}415 of the incorrect tokens,
and copies 5{,}079 incorrect tokens.
Across all supplied proposal positions, the corresponding correct-token retention,
incorrect-token correction, and incorrect-copy rates are 88.69\%, 35.99\%, and 4.96\%.
These unconditional rates also include positions after an earlier draft error.

Table~\ref{tab:app:mechanism-length} gives the sample counts and numerical results for
Figure~\ref{fig:exp:mechanisms}\subref{fig:exp:mechanism-short}.
Every row compares proposal-on and proposal-off predictions at exactly the same anchors.
The positive gains for $L=1,2,3$ support exposing short-match proposals to the learned drafter;
larger gains at longer matches are consistent with the increased specificity of the retrieved continuation.

\begin{table}[htbp]
  \centering
  \caption{Paired proposal-on minus proposal-off gains by match length $L$.
    $\Delta\bar{k}$ is the mean increase in leading reference matches;
    intervals use a conversation-cluster bootstrap.}
  \label{tab:app:mechanism-length}
  \small
  \begin{tabular}{lrrc}
    \toprule
    Match length $L$ & Anchors & $\Delta\bar{k}$ & 95\% CI \\
    \midrule
    1 & 28{,}512 & 0.04 & $[0.02,\,0.05]$ \\
    2 & 10{,}151 & 0.08 & $[0.06,\,0.10]$ \\
    3 & 4{,}763 & 0.16 & $[0.11,\,0.21]$ \\
    4--8 & 7{,}337 & 0.39 & $[0.34,\,0.44]$ \\
    9--22 & 3{,}082 & 1.40 & $[1.23,\,1.59]$ \\
    $\ge 23$ & 3{,}045 & 3.50 & $[3.12,\,3.89]$ \\
    \bottomrule
  \end{tabular}
\end{table}

\FloatBarrier
\section{Main-result evaluation details}
\label{app:main-results}

Each target/temperature configuration in Table~\ref{tab:exp:main} contains 994 requests
from 914 conversations: 128 each for GSM8K, MATH-500, MBPP, LCB, and Alpaca;
30 for AIME25; 164 for HumanEval; and 160 requests from 80 two-turn MT-Bench conversations.
The underlying cases are identical across target sizes and temperatures.
For multi-turn evaluation, each method uses its own previous answer when constructing the next prompt.
At $T=0$, DFlash and \method{} generate identical output token sequences on all 994 requests
for each target size.

For each request, we average the committed lengths of its decoding rounds, counting the
target token, to obtain its acceptance length. Benchmark and Overall values are arithmetic
means over requests. Overall weights requests equally rather than weighting benchmarks or
decoding rounds equally.
Throughput is the sum of generated output tokens divided by the sum of synchronized
per-request wall times, including prefill. Speedup divides this throughput by a fresh AR
measurement on the same requests and under the same execution and sampling settings.
The reported throughput pools individual batch-1 request times.

Table~\ref{tab:app:main-uncertainty} summarizes paired gains over DFlash.
Acceptance-length intervals use 2{,}000 conversation-level bootstrap resamples (seed 42),
keeping the two MT-Bench turns together. These intervals describe evaluation-sample
variation for fixed checkpoints, not variation across training runs or repeated timings.
Each configuration has one complete timed evaluation after two 64-token warm-up runs per method.
The throughput gains are observed ratios from these runs, without a repeated-run timing interval.

\begin{table}[htbp]
  \centering
  \caption{Overall paired gains of \method{} over DFlash in Table~\ref{tab:exp:main}.
    The interval is for the mean per-request acceptance-length difference.
    Relative throughput gains are observed end-to-end TPS ratios.}
  \label{tab:app:main-uncertainty}
  \small
  \setlength{\tabcolsep}{7pt}
  \begin{tabular}{lrrccc}
    \toprule
    Target & $T$ & $\Delta\tau$ & 95\% CI & $\tau$ gain (\%) & TPS gain (\%) \\
    \midrule
    Qwen3-4B & 0 & 0.07 & $[0.04,\,0.09]$ & 1.22 & 0.78 \\
    Qwen3-4B & 1 & 0.05 & $[0.01,\,0.10]$ & 1.08 & 0.67 \\
    Qwen3-8B & 0 & 0.09 & $[0.06,\,0.11]$ & 1.58 & 1.33 \\
    Qwen3-8B & 1 & 0.09 & $[0.05,\,0.14]$ & 2.00 & 0.39 \\
    \bottomrule
  \end{tabular}
\end{table}

\FloatBarrier
\section{Overhead measurement protocol}
\label{app:latency}

\subsection{Transformers measurements}

We isolate the added processing cost of \method{} on identical generation trajectories.
The setup uses Qwen3-8B on an RTX 3090 with BF16 SDPA, $T=0$, batch size 1, block size 16,
and at most 1{,}024 new tokens. We take eight first-turn prompts from each of the eight
benchmarks and run three alternating measurement pairs per prompt.

Both conditions use the same frozen \method{} backbone with mask inputs.
The baseline omits proposal processing. The shadow condition additionally performs index
maintenance, proposal retrieval, input transfer, and reliability-embedding computation, then
discards the proposal-conditioned inputs before the drafter forward pass.
Output tokens and per-round acceptance lengths match in every pair, so the timing difference
measures processing cost without changes in the decoding trajectory.
We use the C++ index and skip empty proposals; copy-feature CUDA graphs are disabled.
Timing synchronizes CUDA at measurement boundaries without adding synchronization between
stages. Model loading and feature-table setup precede measurement.

Decode timing excludes target prefill and initial index construction; end-to-end timing
includes both. We average repetitions within each request and then weight requests equally.
The 95\% intervals in Table~\ref{tab:exp:overhead} use 10{,}000 bootstrap resamples stratified
by benchmark. The total decode increase is computed separately as the ratio of summed decode
times across requests, giving 0.86\%; it is distinct from the 0.42\% mean per-request increase.
Initial index construction takes a mean of 0.355\,ms per request.

\subsection{SGLang measurements}

We run controlled measurements at $T=0$, replicating one HumanEval prompt into fixed batches
of 1, 8, and 16, with six alternating measurement pairs per batch size.
Output tokens and verification-round counts match within each pair.
Table~\ref{tab:exp:overhead-sglang} reports mean extra latency per batch round, including
amortized index initialization, with 95\% bootstrap intervals over the paired measurements.
The wall-time increases are paired medians, rather than the mean per-request increases
reported for Transformers. These fixed-batch measurements isolate overhead on the chosen
prompt; the multi-request serving results are reported separately in
Section~\ref{sec:exp:concurrency}.

\section{Limitations and Future Work}
\label{app:limitations}

\paragraph{A single proposal source.}
Our decoding experiments use only one way to obtain proposals: $n$-gram matching followed
by copying the continuation from the committed context. The effectiveness of \method{}
with other proposal sources remains untested. Future work can explore other ways to
extract useful proposals quickly while keeping their construction cost low.

\paragraph{Modest gains over strong drafters.}
The improvements over the original neural drafters remain modest. Across our experiments,
the additional gains tend to shrink when the baseline drafter is stronger, suggesting that
copied proposals provide less complementary information as draft accuracy improves.
Obtaining larger gains with already strong drafters remains an open challenge.

\paragraph{Benefits concentrated on code.}
The clearest improvements are on code generation. Gains on other tasks are generally
smaller, and performance often remains close to the baseline, particularly in the main
DFlash comparison. This pattern is consistent with the reliance of $n$-gram copying on
useful exact repetitions in the context. Broader benefits may require proposal sources
that capture information beyond exact token matches.


\end{document}